\documentclass[acmsmall]{acmart}
\AtBeginDocument{%
  }

\setcopyright{acmlicensed}
\copyrightyear{2018}
\acmYear{2018}
\acmDOI{XXXXXXX.XXXXXXX}
\acmJournal{CSUR}
\acmISBN{978-1-4503-XXXX-X/2018/06}

\usepackage[draft,commandnameprefix=always,final]{changes}
\usepackage[normalem]{ulem}
\usepackage{xcolor}

\definecolor{addcolor}{RGB}{0,0,200} 
\definecolor{delcolor}{RGB}{200,0,0} 

\setaddedmarkup{\textcolor{addcolor}{#1}}
\setdeletedmarkup{\textcolor{addcolor}{\sout{#1}}}
\usepackage{amsmath}
\usepackage{amsfonts}
\usepackage{booktabs}
\usepackage{graphicx}
\usepackage{makecell}  %
\usepackage{array}
\newcolumntype{P}[1]{>{\raggedright\arraybackslash}p{#1}}
\usepackage{longtable}
\usepackage{booktabs}
\usepackage{tabularx}
\usepackage{caption}
\usepackage{graphicx}
\usepackage{hyperref}
\usepackage{tikz}
\usetikzlibrary{shadows, arrows.meta, positioning}
\usepackage{array}
\newcolumntype{Y}{>{\raggedright\arraybackslash}X}
\usepackage{adjustbox}  
\usepackage{algorithm}
\usepackage{algorithmic}
\usepackage[retainorgcmds]{IEEEtrantools}
\newtheorem{definition}{Definition} 
\usepackage{pifont}
\usepackage{wasysym}
\usepackage[edges]{forest}
\usepackage{tikz}
\usetikzlibrary{shadows} 
\usepackage{xcolor}
\usepackage{graphicx}
\usepackage{graphicx}
\usepackage{lipsum} 
\usepackage{enumitem}
\usepackage{pifont}
\usepackage{xcolor}
\usepackage{tikz}
\usetikzlibrary{arrows.meta, positioning}
\usepackage{makecell}  
\renewcommand{\arraystretch}{1.1}  
\newcommand{\cmark}{{\color{green!60!black}\ding{51}}}   
\newcommand{\xmark}{{\color{red}\ding{55}}}              
\newcommand{\hmark}{{\color{blue}\ding{115}}}          
\newcommand{\wmark}{{\color{orange}\ding{108}}}       
\usepackage{multirow}
\usepackage{color}

\usepackage{booktabs}
\usepackage{multirow}
\usepackage{tabularx}
\usepackage{array}

\newcolumntype{L}[1]{>{\raggedright\arraybackslash}p{#1}}
\newcolumntype{Y}{>{\raggedright\arraybackslash}X}
\usepackage[table]{xcolor} 
\usepackage{lineno}

\begin{document}

\title{Towards Reliable Forgetting: A Survey on Machine Unlearning Verification}

\author{Lulu Xue}
\affiliation{%
  \institution{Huazhong University of Science and Technology}
  \city{Wuhan}
   \state{Hubei}
  \country{China}}
\email{lluxue@hust.edu.cn}
\thanks{This work is supported by the National Natural Science Foundation of China (Grant No. 62372196).}
\author{Shengshan Hu}
\affiliation{%
  \institution{Huazhong University of Science and Technology}
  \city{Wuhan}
     \state{Hubei}
  \country{China}}
\email{hushengshan@hust.edu.cn}
\author{Wei Lu}
\affiliation{%
  \institution{Huazhong University of Science and Technology}
  \city{Wuhan}
     \state{Hubei}
  \country{China}}
\email{luwei_hustcse@hust.edu.cn}

\author{Yan Shen}
\affiliation{%
  \institution{Huazhong University of Science and Technology}
  \city{Wuhan}
     \state{Hubei}
  \country{China}}
\email{yanshen@hust.edu.cn}

\author{Dongxu Li}
\affiliation{%
  \institution{Huazhong University of Science and Technology}
  \city{Wuhan}
     \state{Hubei}
  \country{China}}
\email{dx_li@hust.edu.cn}
\author{Peijin Guo}
\affiliation{%
  \institution{Huazhong University of Science and Technology}
  \city{Wuhan}
     \state{Hubei}
  \country{China}}
\email{gpj@hust.edu.cn}

\author{Ziqi Zhou}
\affiliation{%
  \institution{Huazhong University of Science and Technology}
  \city{Wuhan}
     \state{Hubei}
  \country{China}}
\email{zhouziqi@hust.edu.cn}

\author{Minghui Li}
\affiliation{%
  \institution{Huazhong University of Science and Technology}
  \city{Wuhan}
     \state{Hubei}
  \country{China}}
\email{minghuili@hust.edu.cn}

\author{Yanjun Zhang}
\affiliation{%
  \institution{University of Technology Sydney}
  \city{Sydney}
  \state{NSW}
  \country{Australia}}
\email{Yanjun.Zhang@uts.edu.au}
\author{Leo Yu Zhang}
\affiliation{%
  \institution{Griffith University}
  \city{Brisbane}
  \state{QLD}
  \country{Australia}}
\email{leo.zhang@griffith.edu.au}

\renewcommand{\shortauthors}{Xue et al.}

\begin{abstract}
  With growing demands for privacy, security, and legal compliance (e.g., GDPR), machine unlearning has become a critical technique for ensuring the controllability of learning systems. A central challenge in this area is verifying whether unlearning has been successfully performed. Although unlearning methods are widely studied, verification remains underexplored and lacks a unified framework. This survey addresses the gap by organizing existing methods into behavioral and parametric categories based on the evidence used. It compares representative approaches in terms of assumptions, strengths, and vulnerabilities, and concludes with open problems to support the development of more reliable verification mechanisms.
\end{abstract}




\maketitle
\section{Introduction}
As global concerns over data privacy continue to grow, particularly with the enforcement of regulations such as the General Data Protection Regulation (GDPR)~\cite{regulation2018general}, managing data throughout its lifecycle has become a pressing compliance challenge. In response, machine unlearning has emerged as a key technology for protecting data sovereignty. By selectively removing specific data from trained models, it allows individuals to meaningfully exercise their right to be forgotten by enabling models to forget the influence of that data.
For example, in facial recognition systems, when a user invokes their ``right to be forgotten", traditional forgotten methods merely remove stored records. In contrast, machine unlearning goes further by erasing the data’s influence from the model itself. Similarly, in intelligent customer service systems, when a user requests the deletion of chat history, the unlearning mechanism updates the model to forget the relevant interactions, ensuring that future responses are no longer shaped by the removed content.

Existing machine unlearning techniques can be broadly classified into \textbf{exact unlearning}~\cite{sisa,arcane,e1,e2,e3,e4} and \textbf{approximate unlearning}~\cite{cc,boundary,fast,kd1,kd2}. 
Exact unlearning typically relies on retraining the model to fully eliminate the influence of specific data, which is computationally expensive. To improve the practicality of unlearning in real-world applications, a range of approximate unlearning methods have been proposed. These approaches strike a balance between effectiveness and efficiency, enabling models to approximate forgetting at a significantly lower computational cost.

While algorithmic methods for machine unlearning offer mechanisms to erase specific training data from models, concerns about trust and transparency persist, particularly because most machine learning services operate as opaque, server-side systems. \textbf{A central and unresolved question remains: how can model providers convincingly demonstrate that the targeted data has been thoroughly and irreversibly removed?}

In response, a diverse range of verification techniques has emerged in recent years, drawing from disciplines such as fingerprint~\cite{backdoor1,backdoor2,watermark1,watermark2}, privacy analysis~\cite{mia_game1, mia_game2, mia_game3}, and performance-based diagnostics~\cite{graves2021amnesiac,acc2,bad,zero,towards}. These approaches aim to evaluate whether a model retains residual information after an unlearning operation. However, the absence of a standardized verification framework has led to troubling inconsistencies. For instance, \citet{v_inf} demonstrates that a model may be deemed ``successfully unlearned" under one verification criterion while simultaneously failing another—revealing a disconnect that undermines confidence in current practices.

This lack of alignment raises serious security concerns. Service providers may selectively present verification outcomes to give a false impression of compliance, masking incomplete data removal. In addition, existing verification mechanisms contain security vulnerabilities~\cite{fragile,poul,zhang2024generate,backdoor1,backdoor2} that can be exploited by adversaries to covertly preserve target data while deceiving verification algorithms into confirming successful unlearning.
These issues pose fundamental threats to the security and reliability of machine unlearning.
Therefore, it is essential to establish a unified verification framework and systematically evaluate existing methods. Despite this need, a structured analysis of unlearning verification approaches is still lacking.

\begin{figure*}[h]
    \centering
    \includegraphics[width=0.92\linewidth]{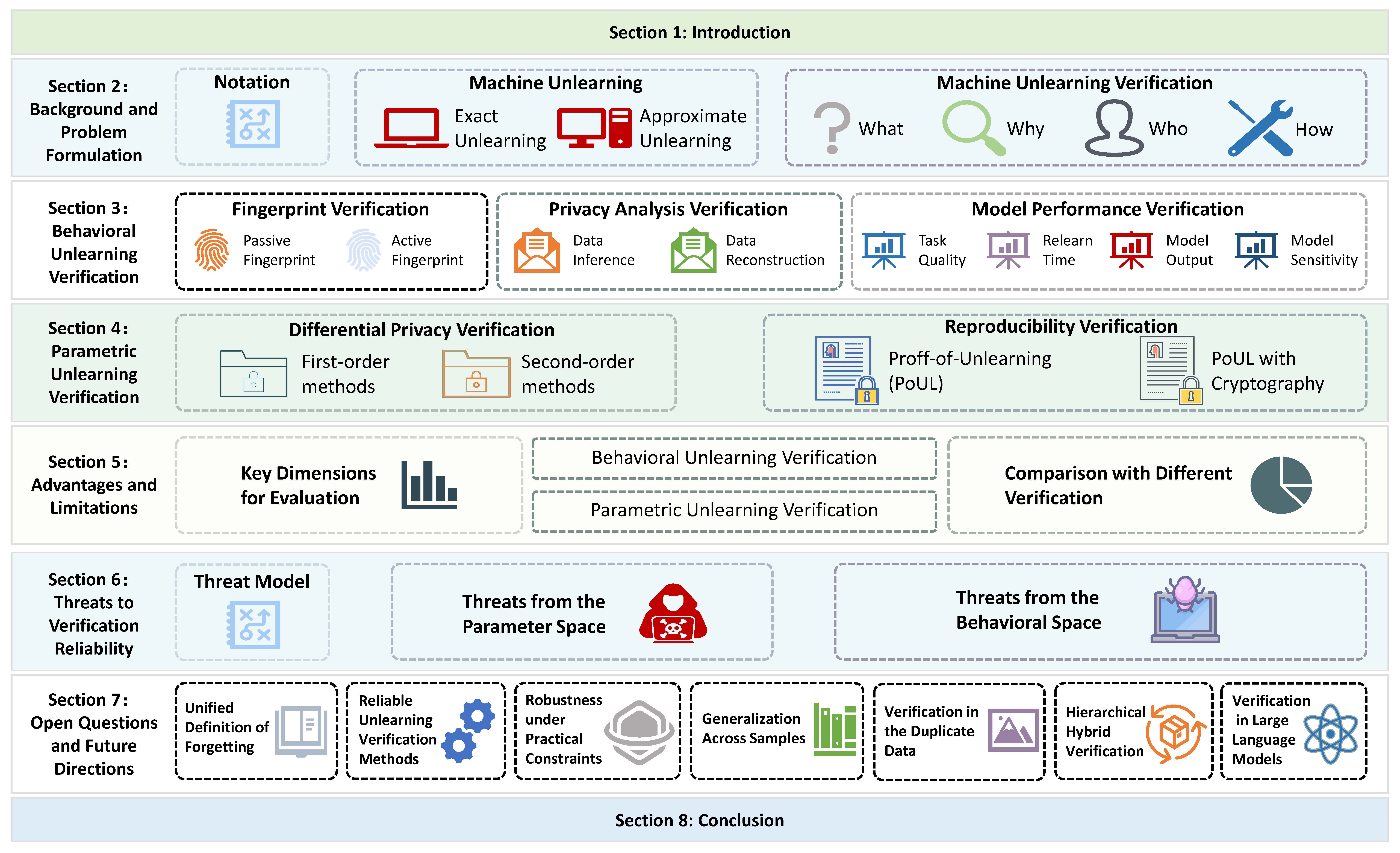}
    \caption{\chadded{An overview of our work.}}
    \label{fig:overview}
\end{figure*}

To address this gap, we present a comprehensive survey of the emerging field of unlearning verification in machine learning. Based on the source of verification signals, we categorize existing approaches into behavioral and parametric methods. Behavioral methods assess whether unlearning has occurred by analyzing the model’s responses to inputs, while parametric methods focus on examining the model’s internal parameters or their derivatives to determine whether the target data has been effectively removed. By structuring our survey around this core distinction, we offer a coherent framework that not only standardizes the terminology used across different methods but also provides a clear lens for comparing their theoretical guarantees, applicability, and robustness. Specifically, our contributions are as follows.

\textbf{Contributions.}
1) 
We present the first structured survey on unlearning verification, categorizing existing methods into behavioral and parametric approaches based on their verification signals. This classification clarifies core assumptions, enables systematic comparison, and highlights key challenges in building reliable verification systems.

2) 
We define seven key evaluation dimensions: theoretical guarantees, access requirements, sample-level verification, verification accuracy, reliance on pre-injected data, efficiency and scalability, and method specificity. Based on these evaluation dimensions, we conduct a detailed comparison to highlight the strengths and limitations of existing verification methods.

3) We further analyze the reliability risks associated with existing methods and outline unresolved issues and future directions, aiming to support the development of verifiable and auditable unlearning systems.

\textbf{{Overview.}}
This paper presents a systematic overview of existing verification techniques for machine unlearning.
The structure of this paper is illustrated in Figure~\ref{fig:overview}. Specifically, 
Section~\ref{sec:background} introduces the formal foundations of unlearning verification, including the definitions of exact and approximate unlearning, the construction of verification predicates, and the roles and interactions involved in the verification workflow.
Section~\ref{sec:empirical} reviews behavioral unlearning verification methods based on observable model behavior, whereas Section~\ref{sec:Formalized} covers parametric approaches that leverage internal model information and provide theoretical guarantees.
Section~\ref{sec:Advantages_limitations} presents a comparative analysis of these approaches, examining their respective strengths and limitations. Section~\ref{sec:threat} investigates the reliability vulnerabilities of unlearning verification under adversarial settings, highlighting how weaknesses in parameter and behavioral signals may be exploited to deceive auditors. Finally, Section~\ref{sec:open} outlines open problems in the field and highlights promising directions for future research.

\begin{table}[h]
  \centering
  \caption{A comparative analysis of various research studies in unlearning methodologies, where MU, FU, GU, LLM-U, and GenAI-U denote Machine Unlearning, Federated Unlearning, Graph Unlearning, Large Language Model Unlearning, and Generative AI Unlearning, respectively. \CIRCLE~ indicates full involvement, \LEFTcircle~indicates partial involvement, and \Circle~ indicates no involvement. It can be observed that existing research methods do not provide a comprehensive investigation into the verification of machine unlearning, and lack sufficient discussion on these methods.}
   \scalebox{0.55}{
    \begin{tabular}{p{4.05em}llllllll}
    \toprule
    {Research} & {Main Area} & 
     \makecell[c]{Fingerprint\\Verification} 
     &  \makecell[c]{Privacy Analysis\\Verification}  & \makecell[c]{Model Performance\\Verification} &  \makecell[c]{Differential Privacy\\Verification}  & \makecell[c]{Reproducibility\\Verification} & \makecell[c]{Vulnerability\\Analysis} & \makecell[c]{Multi-dimensional\\Comparison} \\
    \midrule
    \cite{svy15} & Overview of MU & \Circle    & \Circle    & \Circle    & \LEFTcircle    & \Circle    & \Circle    & \Circle\\
    \cite{svy7} & \multicolumn{1}{p{8.95em}}{{Overview of MU}} & \LEFTcircle    & \LEFTcircle    & \LEFTcircle    & \LEFTcircle    & \Circle    & \LEFTcircle    & \Circle\\
    \cite{svy16} & Overview of MU & \LEFTcircle    & \LEFTcircle    & \LEFTcircle    & \LEFTcircle    &  \LEFTcircle    & \Circle    & \Circle\\
    \cite{svy12} & \multicolumn{1}{p{8.95em}}{{Overview of MU}} & \Circle    & \LEFTcircle    & \LEFTcircle    & \LEFTcircle    & \LEFTcircle    & \Circle    & \Circle\\
    \cite{svy1} & \multicolumn{1}{p{8.95em}}{Overview of MU} & \LEFTcircle    & \LEFTcircle    & \LEFTcircle    & \LEFTcircle    & \LEFTcircle    & \LEFTcircle    & \Circle\\
    \cite{svy4} & Overview of MU & \Circle    & \LEFTcircle    & \LEFTcircle    & \Circle    & \Circle    & \Circle    & \Circle\\
    \cite{svy14} & Overview of MU & \LEFTcircle    & \LEFTcircle    & \LEFTcircle    & \LEFTcircle    & \Circle    & \Circle    & \Circle\\
    \cite{svy8} & \multicolumn{1}{p{8.95em}}{{Overview of MU}} & \LEFTcircle    & \Circle    & \LEFTcircle    & \LEFTcircle    & \Circle    & \Circle    & \Circle\\
    \cite{svy10} & \multicolumn{1}{p{8.95em}}{{Overview of  LLM-U}} & \Circle    & \LEFTcircle    & \LEFTcircle    & \Circle    & \Circle    & \LEFTcircle    & \Circle\\
    \cite{svy17} & Overview of  LLM-U & \Circle    & \Circle    & \LEFTcircle    & \Circle    & \Circle    & \Circle    & \Circle\\
    \cite{svy19} & Overview of FU & \LEFTcircle    & \LEFTcircle    & \LEFTcircle    & \LEFTcircle    &  \LEFTcircle    & \Circle    & \Circle\\
    \cite{svy5} & Overview of FU & \LEFTcircle    & \LEFTcircle    & \LEFTcircle    & \LEFTcircle    & \Circle    & \Circle    & \Circle\\
    \cite{svy3} & Overview of FU & \LEFTcircle    & \LEFTcircle    & \Circle    & \Circle    & \Circle    & \Circle    & \Circle\\

    \cite{svy6} &\multicolumn{1}{p{8em}}{{Overview of  MU and LLM-U}} & \Circle    & \LEFTcircle    & \LEFTcircle    & \Circle    & \Circle    & \LEFTcircle    & \LEFTcircle\\

    \cite{svy9} & {Overview of GU} & \LEFTcircle    & \LEFTcircle    & \LEFTcircle    & \LEFTcircle    & \Circle    & \Circle    & \Circle\\

    \cite{svy13} & Overview of GenAI-U & \Circle    & \LEFTcircle    & \LEFTcircle    & \Circle    & \Circle    & \Circle    & \Circle\\

    \cite{svy11} & Privacy and Security in MU & \Circle    & \Circle    & \Circle    & \Circle    & \Circle    & \Circle    & \Circle\\
    \cite{svy18} & Privacy and Security in MU & \LEFTcircle    & \LEFTcircle    & \Circle    & \Circle    & \Circle    & \Circle    & \Circle\\

    \cite{svy2} & Privacy Risks in MU & \LEFTcircle    & \LEFTcircle    & \LEFTcircle    & \LEFTcircle    & \Circle    & \Circle    & \LEFTcircle\\

    Ours  & Overview of MU Verification & \CIRCLE    & \CIRCLE    & \CIRCLE    & \CIRCLE    & \CIRCLE    & \CIRCLE    & \CIRCLE\\
    \bottomrule
    \end{tabular}%
    }
  \label{tab:comp_study}%
\end{table}%

\textbf{Comparison with Existing Works.}
In recent years, numerous studies have surveyed the landscape of machine unlearning (MU), covering foundational frameworks~\cite{svy1,svy4,svy6,svy7,svy8,svy12,svy14,svy15,svy16}, privacy and security risks~\cite{svy2,svy11,svy18}, federated learning~\cite{svy19,svy3,svy5}, generative models~\cite{svy13}, and large language models~\cite{svy6,svy10,svy17}. However, as shown in Table~\ref{tab:comp_study}, existing surveys generally lack a systematic and dedicated investigation into \textit{unlearning verification}. In most of these works, verification is treated as an auxiliary component of model evaluation rather than as a standalone research focus.
Although some studies touch on verification based on model performance~\cite{svy3,svy7,svy10,svy16}, privacy analysis~\cite{svy2,svy8,svy9,svy11,svy14}, and principles such as differential privacy or reproducibility~\cite{svy3,svy4,svy12}, such discussions are often presented as background or peripheral remarks. They lack a unified framework for systematic categorization or cross-method comparison.
More specifically, these surveys generally do not offer a structured comparison of different verification approaches across core dimensions such as theoretical foundations and evaluation accuracy. They also fail to examine practical considerations in real-world deployments, including feasibility, computational cost, and access constraints. Moreover, potential vulnerabilities and attack surfaces of existing verification techniques remain largely unexplored. These omissions hinder a comprehensive understanding of the capabilities, assumptions, and limitations of current unlearning verification mechanisms.

To bridge these gaps, this work presents the first systematic survey dedicated to verification methods in machine unlearning. We introduce a unified classification framework that encompasses both behavioral and parametric approaches, and assess existing methods across seven key dimensions: theoretical guarantees, access assumptions, sample-level granularity, verification accuracy, reliance on pre-injected data, efficiency and scalability, and method specificity. Our objective is to clarify the applicability and limitations of current verification techniques, and to provide a structured foundation for future research on verifiable machine unlearning in privacy-sensitive and compliance-driven settings.

\section{Background and Problem Formulation}

\label{sec:background}
\subsection{Notation}
To facilitate understanding of the unlearning mechanisms and their verification in this section, the key notation used are summarized in Table~\ref{tab:notation}.

\begin{table}[htbp]
\centering
\caption{Overview of key notation.}
\label{tab:notation}
\scalebox{0.85}{
\begin{tabular}{ll}
\toprule
\textbf{Symbol} & \textbf{Description} \\
\midrule
$A$ & Randomized learning algorithm \\
$D$ & Original training dataset \\
$D_f$ & Forgotten subset of $D$ \\
$D_r$ & Retained dataset, $D \setminus D_f$ \\
$\mathbf{w}^*$ & The model that has never seen $D_f$ \\
$\mathbf{w}$ & Original model \\
$\mathbf{w}'$ & Unlearned model \\
$M$ & Unlearning mechanism \\
$H$ & Hypothesis space induced by $A$ on $D$ \\
$H'$ & Hypothesis space after unlearning \\
$H_r$ & Hypothesis space induced by $A$ on $D_r$ \\
${Q}_{D_f}$ & Query set generator based on $D_f$ \\
$\mathcal{B}$ & Model behavior function \\
$V(\cdot, D_f)$ & Verification predicate for checking unlearning of $D_f$\\
\bottomrule
\end{tabular}
}
\end{table}

\begin{figure*}[h!]
    \centering
    \includegraphics[width=0.5\linewidth]{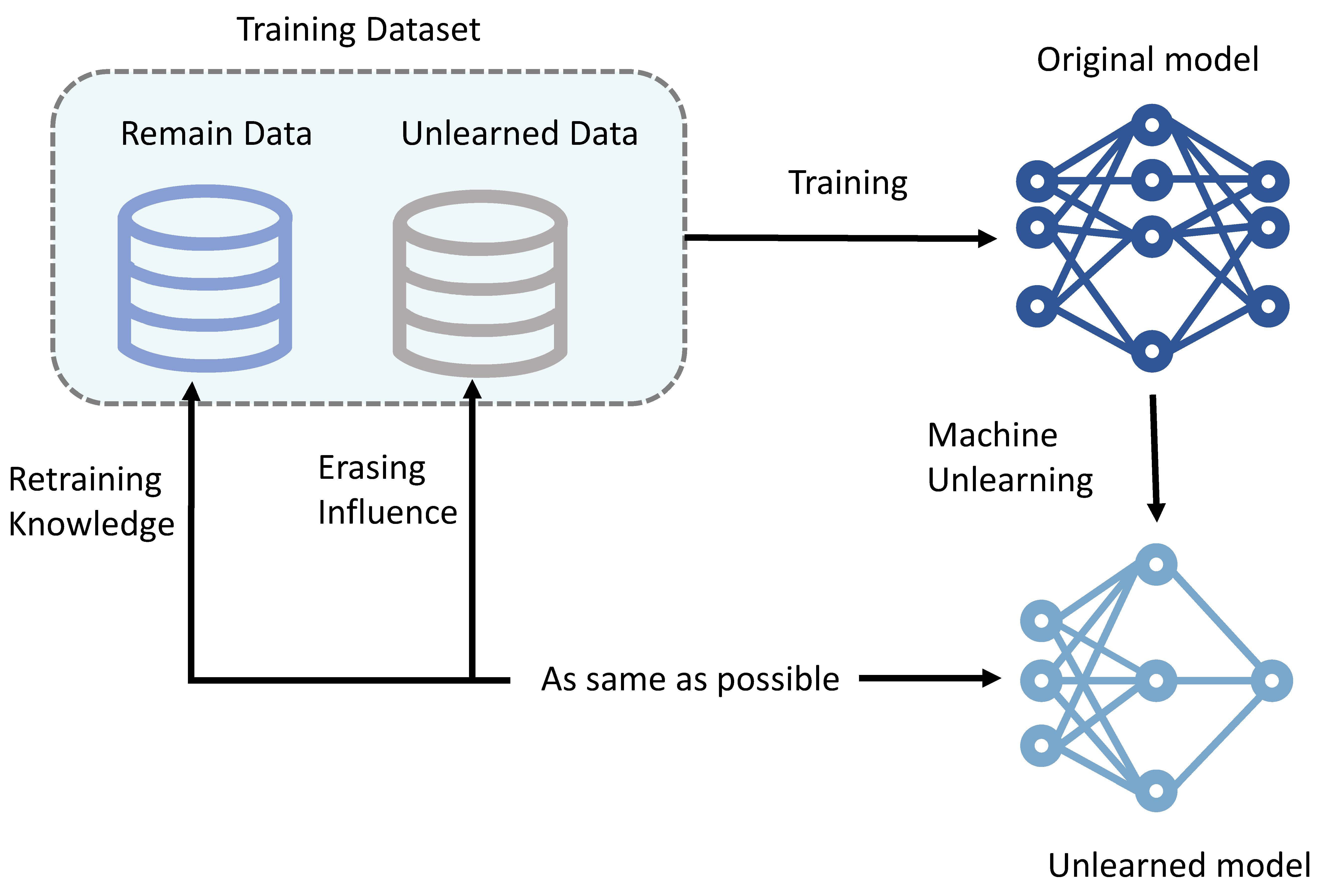}
    \caption{A framework diagram of machine unlearning.}
    \label{fig:mu}
\end{figure*}

\subsection{Machine Unlearning}
\label{sec:mu}
\subsubsection{Definition and Formulation.}

Machine unlearning~\cite{xue2026dual,xue2026unlearnshield} refers to the process of removing the influence of specific training samples from a previously trained machine learning model, while preserving the model’s original performance as much as possible, as shown in Figure~\ref{fig:mu}. Formally, let \( A \) be a randomized learning algorithm which, when trained on a dataset \( D \), induces a distribution over model parameters. This defines a hypothesis space:
\[
H = \text{Range}(A(D)),
\]
which contains all models that can be generated by \( A \) when trained on \( D \).

Let \( \mathbf w \sim A(D) \) denote a specific model sampled from the distribution induced by training on \( D \). We refer to \( \mathbf w \) as the \textit{original model}, as it reflects the full influence of all training samples in \( D \). Suppose a user submits a request to eliminate the influence of a subset \( D_f \subseteq D \), we refer to this subset \( D_f \) as the \textit{forgotten dataset}.

To fulfill such a request, an \textit{unlearning mechanism} \( M \) is introduced. This mechanism operates on the original model \( \mathbf w \) and the forgotten dataset \( D_f \), and produces a unlearned model:
\[
\mathbf w' = M(\mathbf w, D_f), \quad \mathbf w' \in H',
\]
where \( H' \) denotes a new hypothesis space resulting from applying \( M \) to models in \( H \). 

Based on the extent to which the influence of \( D_f \) is removed, existing machine unlearning methods can be broadly classified into two categories: \textbf{exact unlearning} and \textbf{approximate unlearning}.  
\textbf{Exact unlearning} aims to completely forget \( D_f \), requiring that the resulting model be indistinguishable from a model trained from scratch on the retained dataset \( D \setminus D_f \), specifically in terms of its parameters. 
In comparison, \textbf{approximate unlearning} relaxes this requirement by allowing the resulting model to approximate the behavior or parameters of the retrained model. This relaxation enables improvements in computational efficiency, scalability, and real-world deployability. Table~\ref{tab:unlearning-comparison-clean} summarizes representative methods across both categories, which will be discussed in detail below.


\subsubsection{Exact Unlearning}
\label{sec:exact}

{Exact unlearning} refers to the complete elimination of the influence of the forgotten subset \( D_f \) from a trained model. The objective is to produce a model whose distribution is indistinguishable from that induced by retraining on the retained dataset \( D_r = D \setminus D_f \). 

Let \( A \) be a randomized learning algorithm, and let \( M \) be an unlearning mechanism that transforms a specific model \( \mathbf w \sim A(D) \) into a new model \( \mathbf w' = M(w, D_f) \). Exact unlearning is satisfied if the distribution of the unlearned model \( \mathbf w' \) matches that of retraining on \( D_r \), that is,
\[
\mathbf w' \sim A(D_r).
\]
Equivalently, the unlearned model \( \mathbf w' \) is required to lie within the hypothesis space
\[
H_r = \text{Range}(A(D_r)),
\]
which comprises all models attainable when \( A \) is trained solely on the retained data.
This condition ensures that the influence of \( D_f \) is eliminated to the same extent as if those samples had never been included during training.

The most straightforward exact unlearning approach involves full retraining from scratch under the original training setup, often abbreviated as {\textbf{Scratch}}. However, due to its high computational cost, recent work has proposed more efficient exact methods, which can be broadly categorized into \textit{SISA} and \textit{non-SISA} approaches.


\textbf{{SISA Methods.}} 
\citet{sisa} proposed the \textit{Sharded, Isolated, Sliced, and Aggregated (SISA)} training framework, which partitions the training dataset into disjoint shards \( D_1, D_2, \dots, D_K \), each trained independently to produce isolated sub-models \( \mathbf w^1, \mathbf w^2, \dots, \mathbf w^K \). The final model is then obtained by aggregating these sub-models:
\[
\mathbf w = \texttt{Agg}(\mathbf w^1,\mathbf  w^2, \dots,\mathbf  w^K).
\]
When a deletion request is issued for a subset \( D_f \subseteq D_j \), only the affected shard \( D_j \) is updated. The remaining data \( D_j' = D_j \setminus D_f \) is retrained to obtain an updated sub-model \( \mathbf w^{j'} \), and the full model is re-aggregated as:
\[
\mathbf w' = \texttt{Agg}(\mathbf w^1, \dots, \mathbf w^{j'}, \dots, \mathbf w^K).
\]
By localizing the influence of each data point to a specific shard, SISA enables exact unlearning while avoiding full retraining of the entire model.


Subsequent works~\cite{arcane,coded,e1,e2,e3,e4} extend the SISA design to improve unlearning efficiency and model performance. ARCANE~\cite{arcane} partitions the data by class and applies ensemble learning, transforming full-model retraining into per-class slice updates. Aldaghri et al.~\cite{coded} proposed a coding-based sharding protocol, enabling lightweight sub-model recomputation through data-encoded representations. Modular adapter-based retraining~\cite{e1,e2,e3} decouples update-sensitive components from the backbone model, while Chowdhury et al.~\cite{e4} introduced  sequential slice training that enables localized fine-tuning. These designs aim to maintain the semantic guarantee of exact unlearning while minimizing the computational burden.

\textbf{Non-SISA Methods.} Non-SISA methods refer to machine unlearning approaches that do not adopt the "Sharded, Isolated, Sliced, and Aggregated" training paradigm. Specifically, Cao et al. \cite{e5} proposed a method that transforms the learning algorithm into a form dependent on a small number of summations, allowing  training process to be based on these summations rather than individual data points. During the unlearning process, it only requires subtracting the influence of the deleted data from the summations and updating the model. Recently, Liu et al. \cite{e6} introduced a fast retraining method using a distributed Quasi-Newton model update algorithm for data erasure, which is suitable for federated learning scenarios.

\subsubsection{Approximate Unlearning}
\label{sec:Approximate}

Compared to exact unlearning, approximate unlearning relaxes the requirement of strict model equivalence. Specifically, it does not require the unlearned model \( \mathbf w' \) to lie exactly within the hypothesis space \( H_r = \{ A(D_r) \} \). Instead, the unlearning mechanism generates a model that approximates this target in one of the following ways:
\begin{enumerate}
    \item \textbf{Parameter-space approximation}: The unlearned model \( \mathbf w' \) lies within a hypothesis space \( H' \) that is close to \( H_r \) in terms of model parameters.
    \item \textbf{Behavioral-space approximation}: The unlearned model \( \mathbf w' \) behaves similarly to models in \( H_r \) on inputs related to the forgotten dataset \( D_f \), thereby achieving approximate forgetting.
\end{enumerate}

\textbf{Parameter-space approximation.}
Parameter-space approximation aims to ensure that the unlearned model \( \mathbf w' \) remains within a bounded distance from a model obtained via exact retraining.
To achieve such bounded distance, researchers have proposed a range of methods, primarily including first-order approaches based on noise injection or projected gradients, and second-order approaches that utilize Hessian-based corrections.

{\textit{First-order Methods.}}  
First-order methods rely on gradient information to control the shift in model parameters, typically by introducing noise or applying projection techniques to limit changes during model updates.  
For instance, studies such as \cite{langevin, chien2} introduce carefully designed random perturbations in each update step to bound the influence of deleted samples, thereby achieving approximate unlearning in the parameter space.

{\textit{Second-order Methods.}}  
Second-order methods leverage the Hessian matrix or its approximations to more precisely adjust model parameters, aiming to approximate the retraining effect in parameter space.  
A representative example is the Newton Step correction mechanism proposed in \cite{guo}, which updates the model weights using the inverse of the Hessian of the loss function, effectively canceling out the gradient contributions of the deleted data.  
Subsequent works \cite{remember, FedRemoval} build upon this idea by developing Hessian estimation techniques that do not require access to the original training data, extending applicability to scenarios such as federated learning and recommender systems.

\textbf{{Behavioral-space approximation.}}
Behavioral-space approximation refers to ensuring that the unlearned model \( \mathbf w' \) exhibits similar behavior to a retrained model when evaluated on specific inputs, even if their parameters differ.  
Specifically, these methods can be categorized into the following types.

\textit{Fine-tuning (FT)-based Methods.} These methods construct carefully designed datasets and fine-tune the model so that it misclassifies the data intended to be forgotten, making the model behave as if it has never encountered those samples.
Several works~\cite{graves2021amnesiac,cc, boundary} proposed modifying the labels of the data to be forgotten and fine-tuning the model on the resulting modified dataset. This process disrupts the association between the forgotten data and its correct label, allowing the model to gradually forget the data that no longer needs to be retained. Recently, to enhance privacy protection for forgotten data, some studies explore unlearning methods that do not require access to the forgotten data. \citet{fast} implemented unlearning by fine-tuning using only a subset of the retained dataset and incorporating a specially constructed noise matrix. \citet{zero} extended the method from \cite{fast} to apply to scenarios without the original data, further improving privacy protection.

\textit{Gradient Ascent (GA)-based Methods.}  
These methods, as demonstrated by \cite{deltagrad, BAERASER}, apply gradient ascent to the loss computed on the data to be forgotten.  Recently, \citet{kim2025rethinking} leverages the idea of gradient ascent to design an unlearning verification method tailored for text generation tasks in large multimodal models. This process degrades the model's performance on that data, thereby driving its behavior closer to that of a model that has never encountered the forgotten data, and ultimately achieving approximate unlearning in the behavioral space.

\textit{Influence-based Methods.} 
These methods aim to assess and eliminate the influence of specific training samples on model predictions or outputs. By leveraging influence functions or the Fisher Information Matrix (FIM) to quantify the impact of individual data points on models, these methods enable fine-grained control over the retention of training information.
\citet{i6} proposed a weight-sanitization approach that adjusts model parameters using the fisher information matrix.
\citet{i4} introduced a neural tangent kernel linearization technique to more accurately estimate the direction of influence, and combine it with noise injection to mitigate residual effects.
Building on influence function theory, \citet{ssse} presented SSSE, a practical and scalable unlearning method. This method approximates the Hessian matrix using an empirical FIM and uses the Sherman-Morrison formula to efficiently compute its inverse, allowing the removal of the deleted sample’s influence without requiring access to the full training set.
It is important to emphasize that these methods are fundamentally different from parameter-space approximation approaches, as they aim to eliminate the influence on model behavior rather than constrain differences in parameter space.

\textit{{Knowledge Distillation (KD)-based Methods.}}
Knowledge distillation~\cite{kd1,kd2} is a technique that improves a student model by transferring behavioral patterns or soft targets from a more complex teacher model. 
In the context of machine unlearning, these methods treat the student as the unlearned model and achieve forgetting by blocking the transfer of behavior patterns from the teacher that are associated with the data to be forgotten.
\citet{zero} introduced a band-pass filter to remove data containing forgotten information, ensuring that the student model only receives the retained information. \citet{bad} established a teacher-student framework with both competent and incompetent teachers, using selective knowledge transfer to enable the student model to forget certain information. \citet{towards} presented the SCRUB method, which effectively achieves forgetting by minimizing KL divergence and task loss on non-forgotten datasets while maximizing KL divergence on forgotten datasets.

\textit{Membership Game-based Methods.}  
These methods focus on aligning membership-related behavior, aiming to ensure that the final model exhibits membership inference responses on the forgotten dataset \( D_f \) that are indistinguishable from those of a model that has never encountered \( D_f \). In doing so, they seek to eliminate observable behavioral traces of the forgotten data.
Recent works~\cite{mia_game1, mia_game2, mia_game3} formalized this process as a game between the unlearning mechanism and an adversary, where the model's behavior is optimized to evade membership inference attacks. However, these methods rely exclusively on behavioral signals related to membership privacy over the forgotten dataset, and their effectiveness heavily depends on the strength of the specific attack strategy used. As a result, their reliability and generalizability are highly sensitive to the choice and quality of the underlying inference mechanism.

\begin{table}[htbp]
\centering
\footnotesize  
\caption{Comparison of unlearning methods. \textit{Approx.~(Param.)} denotes approximate unlearning based on parameter-space approximation, while \textit{Approx.~(Behavior)} refers to behavioral-space approximation.}
\renewcommand{\arraystretch}{1.1}
\scalebox{0.95}{
\begin{tabularx}{\textwidth}{
  >{\raggedright\arraybackslash}p{3cm}
  >{\raggedright\arraybackslash}X
  >{\raggedright\arraybackslash}X
  >{\raggedright\arraybackslash}X
}
\toprule
\textbf{Aspect} & \textbf{Exact Unlearning} & \textbf{Approx.~(Param.)} & \textbf{Approx.~(Behavior)} \\
\midrule

\textbf{Goal} &
Achieve a model unexposed to the forgotten data. &
Similar in parameter space to a model unexposed to the forgotten data. &
Similar in behavioral space to a model unexposed to the forgotten data.\\
\midrule
\textbf{Methods} & Scratch, SISA, Non-SISA & First-order, Second-order & FT, GA,  Influence, KD, Membership games\\

\midrule
\textbf{Parameter Equivalence} & Yes & Approximate & N/A \\
\textbf{Behavioral Equivalence} & Yes & N/A & Approximate \\
\bottomrule
\end{tabularx}
}
\label{tab:unlearning-comparison-clean}
\end{table}

\subsection{Machine Unlearning Verification}
\label{sec:uv}

\begin{figure*}[h!]
    \centering
    \includegraphics[width=0.9\linewidth]{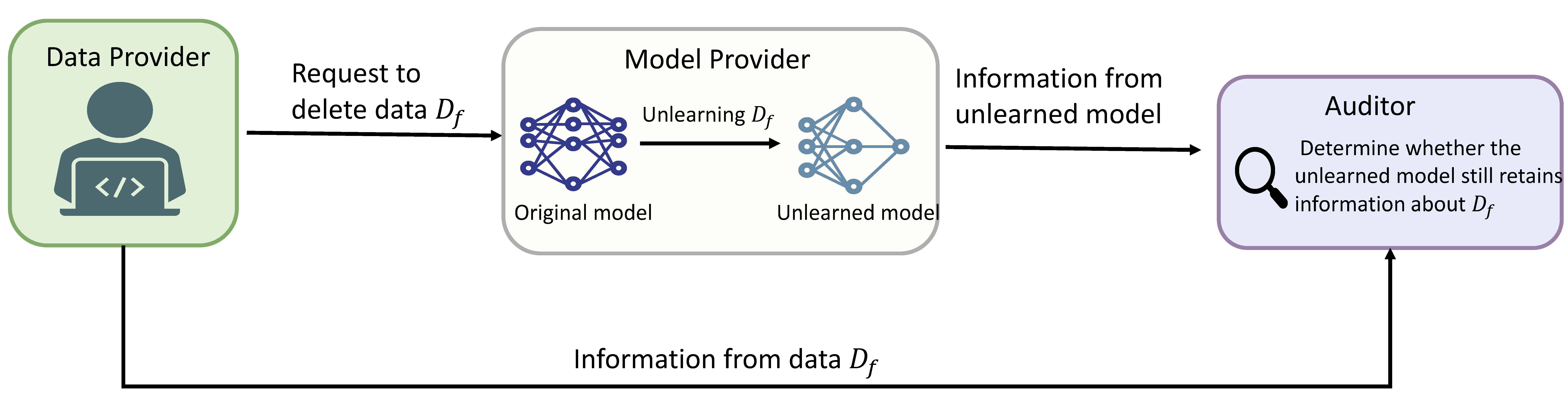}
    \caption{The process of machine unlearning verification.}
    \label{fig:uv}
\end{figure*}

We structure our discussion of machine unlearning verification methods around four guiding questions: \textbf{What}, \textbf{Why}, \textbf{Who}, and \textbf{How}. These questions collectively outline the fundamental form, motivation, key participants, and technical landscape of unlearning verification.

\textbf{What is unlearning verification?}
An {unlearning verification method} aims to determine whether a machine learning model has successfully forgotten a specified subset of its training data.
To formalize this notion, we express unlearning verification as the evaluation of a binary predicate:
\[
V(\mathbf w', D_f) \in \{0,1\},
\]
where \( V(\mathbf w', D_f) = 1 \) indicates that unlearning has succeeded, meaning no residual influence of \( D_f \) is detectable in the model \( \mathbf w' \); whereas \( V(\mathbf w', D_f) = 0 \) denotes that the unlearning process has failed to fully remove the impact of \( D_f \).

    \textbf{{Why is unlearning verification necessary?}}
The necessity of unlearning verification arises from both regulatory and methodological imperatives.
From the regulatory and trust perspective, unlearning verification provides the foundation for accountability in machine learning systems. As models increasingly incorporate sensitive or legally protected data such as medical records, financial transactions, or user activity logs, the ability to verifiably erase such data becomes not only a best practice but also a legal requirement. Regulatory frameworks such as the GDPR explicitly mandate the right to be forgotten. However, in the absence of a trustworthy verification mechanism, any claim of compliance remains unsubstantiated. Unlearning verification addresses this gap by enabling model providers to generate verifiable compliance artifacts, while allowing users and auditors to independently assess whether data deletion has genuinely occurred.

From the methodological perspective, unlearning verification is essential for assessing the soundness and reliability of approximate unlearning. As discussed in Section~\ref{sec:Approximate}, approximate unlearning, in contrast to exact unlearning, involves deliberate trade-offs among performance, efficiency, and feasibility. It typically aims for the model’s parameters or behavior to converge toward the state of a model that has never been exposed to the forgotten data. Under this relaxed semantic assumption, unlearning verification serves as a critical tool for quantifying the extent of forgetting, helping to answer the question of whether the model has sufficiently forgotten the data. Thus, unlearning verification is not only a supporting mechanism for regulatory compliance, but also an indispensable component in the design and evaluation of unlearning methods.

\textbf{Who participates in the verification process?}
The unlearning verification process typically involves three main entities: the \textit{data owner}, who requests the removal of a specified subset \(D_f\) from the training data; the \textit{model provider}, who applies an unlearning mechanism and returns a model \(\mathbf w'\) that is claimed to have forgotten \(D_f\); and the \textit{auditor}, who evaluates whether \(\mathbf w'\) still retains information about the forgotten data, as illustrated in Figure~\ref{fig:uv}.

While these roles are conceptually distinct, in practice they may overlap depending on the deployment context. For instance, in user-centric settings, the data owner may also act as the auditor. In contrast, in scenarios involving outsourced training or institutional oversight, verification is often delegated to an independent third party responsible for assessing compliance on behalf of users or regulatory authorities.

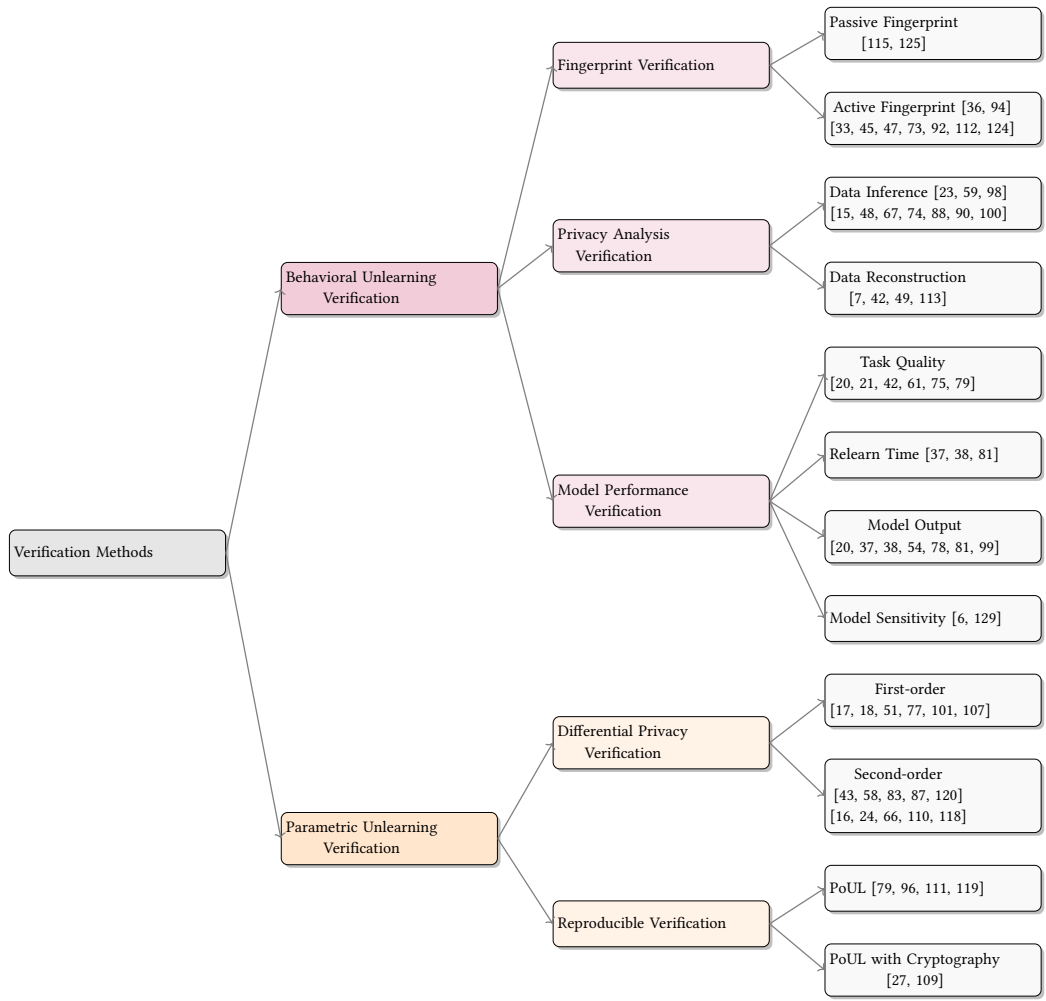
\begin{figure*}[t!]
\centering
\scalebox{0.6}{
\begin{forest}
for tree={
  font=\large,
  grow'=0,                         
  draw,                            
  rounded corners,
  align=center,
  parent anchor=east,
  child anchor=west,
  l sep=1.2cm,                     
  s sep=0.7cm,                     
  edge={->, thick, gray},
  minimum height=1.0cm,
  text width=4.5cm,
  fill=gray!5,
  drop shadow
}
[Verification Methods, fill=gray!20
  [Behavioral Unlearning\\
Verification, fill=purple!20
    [Fingerprint Verification, fill=purple!10
      [Passive Fingerprint~\\\cite{xuan2025verifying, zhang2024generate}]
      [Active Fingerprint \cite{goel2022towards, tam2024towards}\\\cite{gao2024verifi,guo2023verifying,han2025vertical, zhang2024duplexguard, sommer2022athena, xu2024really, lu2025waterdrum}]
    ]
    [Privacy Analysis\\ Verification, fill=purple!10
      [Data Inference~\cite{mia_game1, mia_game2,kim2025rethinking}\\\cite{mia_game3,liu2023muter,seo2025revisiting,chen2021machine,hayes2024inexact,lu2022fp,wang2024towards}]
      [Data Reconstruction\\\cite{graves2021amnesiac,xu2024evaluating,bertran2024reconstruction,hu2024learn}]
    ]
    [Model Performance\\ Verification, fill=purple!10
      [Task Quality\\\cite{graves2021amnesiac,acc2,bad,zero,towards,nguyen2022markov} ]
      [Relearn Time~\cite{i4,i6,ssse}]
       [Model Output\\ \cite{i4,i6,bad,nguyen2020variational,ssse,v_inf,v_ai} ]
      [Model Sensitivity \cite{zhou2025truvrf,v_fish}]
    ]
  ]
  [Parametric
Unlearning\\
Verification, fill=orange!20
    [Differential Privacy \\ Verification, fill=orange!10
      [First-order\\\cite{langevin, chien2,vfl,vfr, vword, rewind} ]
      [Second-order\\\cite{guo, remember,FedRemoval, second2,hessianfree}\\\cite{minmax, efficient_graph, graph_2,certified_edge, scalable}]
    ]
    [Reproducible Verification, fill=orange!10
      [PoUL~\cite{poul,fragile, deltagrad, nguyen2022markov}]
      [PoUL with Cryptography\\\cite{tee,crypt}]
    ]
  ]
]
\end{forest}
}
\caption{A taxonomy of unlearning verification methods.}
\label{fig:forest_purple_orange}
\end{figure*}

\textbf{{How to perform unlearning verification?}}
The core objective of unlearning verification is to determine whether a model has successfully removed the influence of a specified data subset. As discussed in Section~\ref{sec:mu}, the purpose of existing unlearning methods is to align the model’s parameters or behavior with those of a model that has never seen the forgotten data, either approximately or exactly.   Accordingly, we categorize unlearning verification methods into two types: {behavioral
unlearning verification} and parametric unlearning verification, which evaluate whether the forgotten data remains in the model from the perspective of external behavior or internal parameters, respectively.

\begin{enumerate}
    \item \textbf{{Behavioral
unlearning verification.} }
Behavioral unlearning verification aims to determine whether a model has successfully forgotten the $D_f$ by evaluating its responses to specific inputs.  
Such methods typically define behavioral characteristics of the model and assess a predicate \( V(\mathbf w', D_f) \) based on observable changes in these behaviors.  
Specifically, the evaluation focuses on the model’s behavior \( \mathcal{B}(\mathbf w', Q_{D_f}) \) over a designated query set \( Q_{D_f} \), which contains inputs related to the forgotten data \( D_f \). 
If this behavior is consistent with or sufficiently close to that of \( \mathcal{B}(\mathbf w^*, Q_{D_f}) \), where \( \mathbf w^* \) is a model that has never seen \( D_f \), the unlearned model is considered to have successfully forgotten \( D_f \).
\item \textbf{{Parametric unlearning verification.}} 
These methods evaluate whether unlearning has been achieved by measuring the difference in parameter space between the target model and a reference model that has never been exposed to the forgotten data. If the difference falls below a predefined threshold, the model is considered to meet the unlearning criterion. These approaches rely on specific formal assumptions, as discussed in Section~\ref{sec:Formalized}. Compared to behavioral verification, parametric verification typically offers greater reliability. However, it often requires access to internal model parameters or training dynamics, and may incur higher computational costs. 
\end{enumerate}

These two paradigms differ substantially in terms of verification strength, computational efficiency, and deployment feasibility. Figure~\ref{fig:forest_purple_orange} provides an overview of this taxonomy and serves as a reference framework for the analysis in the following sections. \chadded{
Table~\ref{tab:verification_metrics_functions} summarizes the  quantitative metric with each category of methods, which is beneficial for conducting unified benchmarking in future work.
}
\begin{table}[htbp]
\centering
{
\caption{Quantitative Metrics for Unlearning Verification}
\label{tab:verification_metrics_functions}
\renewcommand{\arraystretch}{1.18}
\setlength{\tabcolsep}{4pt}

\scalebox{0.87}{
\begin{tabularx}{\textwidth}{L{3.1cm} L{3.1 cm} L{3.3cm} Y}
\toprule

\textbf{Category} & \textbf{Method} & \textbf{Metric} & \textbf{Related Mathematical Functions} \\
\midrule

\multirow{2}{=}{Fingerprint Verification}
& Passive Fingerprint
& \multirow{2}{=}{Fingerprint Detection Rate}
& \multirow{2}{=}{N/A} \\
& Active Fingerprint
& & \\
\midrule

\multirow{2}{=}{Privacy Analysis Verification}
& Data Inference
& Inference Success Rate
& N/A \\
& Data Reconstruction
& Data Reconstruction Quality
& N/A \\
\midrule

\multirow{7}{=}{Model Performance Verification}
& Task Quality
& Task Performance
& N/A \\
& Relearn Time
& Relearning Time
& N/A \\

\cmidrule(lr){2-4}

& \multirow{4}{=}{Model Output}
& Interpretability of Outputs
& N/A \\
&
& Confusion Matrix
& N/A \\
&
& Output Distance
& KL Divergence, L2 Norm \\
&
& Mutual Information of Outputs
& Mutual Information \\

\cmidrule(lr){2-4}

& Model Sensitivity
& The Sensitivity of $w'$
& Fisher Information, Loss Gradient \\
\midrule

\multirow{2}{=}{Differential Privacy Verification}
& First-order
& {Indistinguishability of}
& \multirow{2}{=}{Differential privacy} \\
& Second-order
&  Parameter Distributions& \\
\midrule

\multirow{2}{=}{Reproducible Verification}
& PoUL
& \multirow{2}{=}{Parameter Distance}
& \multirow{2}{=}{L2 Norm, Frobenius Norm} \\
& PoUL with Cryptography
& & \\
\bottomrule
\end{tabularx}
}
}
\arrayrulecolor{black}
\normalcolor
\end{table}

\section{Behavioral Unlearning Verification}
\label{sec:empirical}
 Existing behavioral verification methods can be classified into three categories: fingerprint methods, privacy analysis methods, and model performance methods. These approaches differ in the specific types of model behaviors they exploit as verification signals, as well as in the intuitions underlying those behaviors. 
  Their behavioral constructions are summarized in Table~\ref{tab:behavior-definitions}.


\begin{table}[h!]
\centering
\caption{Behavioral definitions across behavioral verification methods.}
\label{tab:behavior-definitions}
\scalebox{0.72}{
\begin{tabular}{lccc}
\toprule
\textbf{Aspect} & \textbf{Fingerprint} & \textbf{Privacy Analysis} & \textbf{Model Performance} \\
\midrule
\textbf{Behavior} & 
\makecell[l]{Detection of \\ fingerprint} & 
\makecell[l]{Retention of \\ privacy information} & 
\makecell[l]{Model performance} \\
\midrule
\makecell[l]{\textbf{Intuition}} & 
\makecell[l]{Models not trained on \( D_f \) \\ do not show its fingerprints} & 
\makecell[l]{Models not trained on \( D_f \) \\ should not leak its privacy} & 
\makecell[l]{Models trained with and without \\ \( D_f \) perform differently on \( D_f \)} \\
\midrule
\textbf{Types of Behavior} & 
\makecell[l]{Passive Fingerprint, \\Active Fingerprint} & 
\makecell[l]{Data Inference, \\ Data Reconstruction} & 
\makecell[l]{Task Quality, Relearn Time, \\ Model Output, \\ Model Sensitivity} \\
\bottomrule
\end{tabular}
}
\end{table}

\subsection{Fingerprint}
In machine learning, \emph{fingerprints} refer to distinctive response patterns a model acquires from exposure to specific training data~\cite{f1,f2}. These patterns can be detected through a query set \( Q_{D_f} \), which probes whether the model retains behavioral traces associated with the forgotten data \( D_f \). The model’s responses \( \mathcal{B} \) reveal whether such residual fingerprints remain.\chreplaced{}{
The core intuition is that models trained on \( D_f \) often exhibit systematic deviations from those never exposed to it, for example, consistently misclassifying triggered inputs into a specific class or responding differently to adversarially perturbed inputs. Fingerprint verification audits such signals to assess whether \( D_f \) has been truly forgotten.}
Fingerprint verification methods can be categorized as active or passive, based on whether the unlearned data has undergone prior proactive intervention.

\subsubsection{Passive Fingerprint.}
Passive fingerprint verification does not rely on any modification to the training data. Instead, it infers the presence of forgotten information by observing differences in the model's behavior under natural conditions or minimally perturbed inputs. A common strategy in this category is to use adversarial examples~\cite{wang2025advedm, song2025segtrans, song2026erosion, zhou2025vanish} as post hoc probes to detect residual responses in the model parameters, thereby assessing whether the unlearning process has been truly effective.

Adversarial example techniques refer to methods for generating inputs with imperceptible perturbations that intentionally trigger incorrect or unstable outputs in machine learning models~\cite{zhou2024securely,zhou2023downstream,zhou2023advclip,li2024transfer,zhou2025numbod,zhou2024darksam,wang2025breaking,song2025segment}. Although originally developed to test model robustness and to design attacks, these techniques have increasingly been repurposed for auxiliary objectives such as unlearning verification. By using adversarially perturbed inputs as diagnostic probes, researchers can assess whether a model remains sensitive to specific data points that are supposed to have been erased.

\citet{xuan2025verifying} proposed a white-box adversarial verification method for machine unlearning, termed Unlearning Mapping Attack (UMA), which is applicable to both classification and generative models. In classification tasks, UMA perturbs known forget samples to check whether the model still produces the same predictions as before unlearning. If it does, this suggests that the model retains latent representations associated with the forgotten class, indicating incomplete unlearning. This residual sensitivity implies that decision boundaries related to the forgotten data have not been effectively removed. In generative tasks, UMA crafts adversarial perturbations on the masked inputs (i.e., partially occluded versions) of samples from the forget set. The goal is to induce the unlearned generative model to produce outputs that closely resemble those generated by the original model. Successful reconstruction indicates that the model still retains residual information about the forgotten data.

Similarly, \citet{zhang2024generate} proposed UnlearnDiffAtk, which constructs adversarial prompts specifically for diffusion models. If the unlearned model still generates forgotten content in response to these prompts, it is considered to have failed the unlearning verification.

\subsubsection{Active Fingerprint.}
Active fingerprint verification involves constructing or modifying specific data samples related to the unlearning target before or during training, in order to embed identifiable fingerprint information.
Several active fingerprint methods construct verifiable fingerprints for unlearning verification by strategically modifying the target data to be forgotten, such that only models exposed to the data can correctly identify them. For instance, Goel et al.~\cite{goel2022towards} proposed {Inter-class Confusion Testing}, which generates adversarially perturbed samples during training and later evaluates the model's ability to recognize these samples after unlearning. If the model fails to recognize them, the unlearning is considered successful. In the context of federated learning, Tam et al.~\cite{tam2024towards} introduced adversarial noise as validation markers to assess whether information from a specific domain has been effectively removed. Similarly, Gao et al.~\cite{gao2024verifi} explored the use of label-flipping techniques to construct verification signals for unlearning.

In addition, several studies have demonstrated the feasibility of using backdoor mechanisms for unlearning verification. Essentially, a backdoor attack is a form of data poisoning in which the attacker injects trigger-embedded samples into the training data~\cite{guo2023verifying,hu2022badhash,wang2024trojanrobot,zhang2024detector,zhang2025test}. Once backdoored, the model behaves normally on clean data but produces targeted incorrect predictions when presented with inputs containing the trigger. This attack paradigm can be repurposed for unlearning verification: the model owner deliberately implants backdoor triggers into the model, and after executing unlearning algorithms, checks whether the model’s predictions on the trigger inputs have changed. If they remain the same, unlearning is deemed to have failed; otherwise, it is considered successful.
Recently, studies~\cite{guo2023verifying,han2025vertical} have proposed backdoor-based verification strategies, in which specific triggers are embedded into target data during training. In the verification phase, the model's response to these trigger-embedded queries is monitored to determine whether it still predicts them as the user-specified target class. This approach enables black-box unlearning verification without requiring access to the model's internal parameters. \citet{zhang2024duplexguard} proposed {DuplexGuard}, which associates each forget subset with a corresponding emergent subset. The model exhibits no backdoor behavior before unlearning, but once the forget subset is removed, the backdoor effect on the emergent subset becomes active, serving as a verification signal.
In study~\cite{sommer2022athena}, users inject unique backdoor triggers into their own data, and the model's response to these triggers after deletion requests is statistically evaluated using hypothesis testing. This provides a quantitative and high-confidence verification mechanism that remains robust against state-of-the-art backdoor defenses and adaptive server behaviors. \citet{xu2024really} proposed IndirectVerify, a novel unlearning verification scheme that constructs influential sample pairs, consisting of a trigger sample and a corresponding reaction sample. By perturbing the trigger sample to influence the classification of the reaction sample via gradient matching, the method ensures that successful unlearning modifies the prediction of the reaction sample.

In addition to backdoor-based approaches, \citet{izzo2021approximate} proposed a \textit{feature injection test} to evaluate the effectiveness of unlearning. This method introduces an artificial feature into \(D_f\) that is activated only for a small subset of samples and is highly predictive of the label. During training, the model typically develops a strong response to this feature. Therefore, if the unlearning mechanism is effective, removing the associated data should lead to a significant reduction in the model’s reliance on this feature, which can manifest as changes in predictive behavior. \citet{lu2025waterdrum} proposed WaterDrum, a black-box unlearning verification method for large language models that adopts an active fingerprinting paradigm. It embeds robust text watermarks, parameterized by private cryptographic keys, into the training data. Models trained on such data emit outputs containing identifiable signals, which vanish once the data is successfully unlearned. Verification is conducted privately by the data owner using their key to detect the presence or absence of the watermark.

\subsection{Privacy Analysis}
Privacy analysis verification evaluates the effectiveness of unlearning by constructing a privacy analysis module that examines the leakage of sensitive information related to \( D_f \). In this approach, the query set \( Q_{D_f} \) typically consists of data points sampled directly from \( D_f \), and the behavioral function \( \mathcal{B} \) measures the model’s responses under the privacy analysis module\chadded{.}\chdeleted{, reflecting its potential retention of information about \( D_f \).}  
\chdeleted{The core intuition is that a model which has not forgotten \( D_f \) and a model \( \mathbf w^* \) that has never seen \( D_f \) should exhibit significantly different privacy signals when evaluated through the privacy analysis module on \( Q_{D_f} \).  
For example, when using membership inference for verification, one can infer that \( D_f \) belongs to the training set from the non-unlearned model, whereas \( D_f \) would be inferred as a non-member in the model \( \mathbf w^* \).} This type of verification primarily utilizes the following privacy analysis techniques: data inference and and data reconstruction.


\subsubsection{Data Inference}
Existing data inference approaches for unlearning verification can be broadly categorized into two types: Membership Inference (MI) and Non-Membership Inference (NMI). MI methods aim to determine whether the target model has encountered a specific data point during training, and have been widely used to distinguish between training and unseen data~\cite{chen2021machine, shokri2017membership}. Some studies further adapted MI to the unlearning verification setting, where a data point is considered successfully unlearned if it is inferred as a non-member after the unlearning request~\cite{mia_game1, mia_game2, mia_game3,liu2023muter,seo2025revisiting}. Additionally, \citet{kim2025rethinking} proposed an implicit membership inference framework tailored for large vision-language models in text generation tasks, aimed at evaluating whether the model has truly forgotten sensitive information related to specific entities. In this approach, Wikipedia-style summaries of target entities are used as proxies for private information. By measuring the semantic similarity between the model’s generated output and the entity summary, the method indirectly assesses whether memorization persists. Specifically, if the output shows significantly higher similarity to the target entity’s description than to that of unrelated entities, it is considered indicative of residual membership bias, suggesting incomplete forgetting. This method does not require access to model parameters and is applicable in black-box settings.

%

MI focuses on identifying which data the unlearned model still retains, relying solely on access to the unlearned model. In contrast, NMI aims to identify which data points have been forgotten by comparing model responses before and after unlearning, thus requiring access to both the original and unlearned models.
Some works~\cite{chen2021machine,hayes2024inexact} have proposed inference methods that leverage discrepancies between the outputs of the original and unlearned models. Lu et al.~\cite{lu2022fp} extended  this idea to label-only black-box settings, which infer unlearning success based on the amount of perturbation required to change predicted labels, even when only hard labels are accessible. Further, \citet{wang2024towards} systematically introduced  NMI into the unlearning verification framework, utilizing model state transitions to enhance the reliability of unlearning validation. To enable more efficient verification, their method operates in a white-box setting and directly performs inference on the target model, eliminating the need to train multiple shadow models.

\subsubsection{Data Reconstruction}  
Data reconstruction is a class of privacy analysis techniques that aim to recover specific features of target data from a given model~\cite{fredrikson2015model, buzaglo2023deconstructing, haim2022reconstructing}. Some studies~\cite{graves2021amnesiac,xu2024evaluating} show that such techniques can be used to evaluate whether a model successfully forgets particular data. Graves et al.~\cite{graves2021amnesiac} demonstrated that class-level reconstruction reveals whether a model retains the semantic information of certain categories; however, it lacks the granularity needed to verify forgetting at the sample level. To address this limitation, Xu et al.~\cite{xu2024evaluating} proposed a parameter-driven reconstruction framework that formulates an optimization objective based on implicit bias theory. Their method enables direct recovery of previously seen training samples from model parameters, without requiring labels or external triggers, making it well-suited for general-purpose unlearning verification.

The core idea behind these approaches is that a properly unlearned model should no longer be able to reconstruct information associated with the forgotten data. If such information remains recoverable, it indicates that the unlearning process is incomplete. Recently, some works~\cite{bertran2024reconstruction,hu2024learn} have introduced a more direct verification approach: attempting to reconstruct the specific data samples that have been requested for deletion. If a reconstructed sample matches one of these forgotten targets, the data is considered not effectively forgotten. \citet{bertran2024reconstruction} implemented reconstruction-based verification for unlearned samples in simple models; however, their approach does not scale well to complex deep neural networks. To address this limitation, \citet{hu2024learn} applied gradient inversion techniques to enable the reconstruction of unlearned samples in more complex architectures, demonstrating improved effectiveness in realistic settings.

\subsection{Model Performance}
Model performance unlearning verification methods aim to evaluate the effectiveness of unlearning by assessing the model's performance on forgotten data. In this method, the query set \( {Q}_{D_f}\) consists of data points from \( D_f \). The behavior \( \mathcal{B} \) represents the model's performance on the query set \( {Q}_{D_f}\). 
\chdeleted{The core intuition behind such methods is that models exposed to \( D_f \) will exhibit different performance compared to models that have never seen \( D_f \).}
Existing performance-based unlearning verification metrics include task quality, retraining time, model outputs, and model sensitivity.


\subsubsection{Task Quality.}
Task quality refers to a model’s ability to perform its intended function on a given dataset. This is typically measured by performance metrics specific to the task, such as classification accuracy for supervised learning, semantic coherence for text generation, or perceptual quality for image synthesis.
Task quality evaluation on the forgotten dataset is a widely used and intuitive approach for assessing unlearning effectiveness~\cite{graves2021amnesiac,acc2,bad,zero,towards}. The underlying idea is that if the model’s task quality on the forgotten data significantly decreases, this indicates that the model has reduced or lost its reliance on those samples, suggesting that forgetting has occurred. 
In addition, \citet{nguyen2022markov} proposed  comparing the task quality of the unlearned model with that of a model retrained from scratch on the retained dataset\chreplaced{.}{, which has never seen the erased data.} Although this method provides a useful benchmark, it also introduces additional computational overhead due to the need for retraining. Recently, \citet{kim2025rethinking} evaluated forgetting by measuring task quality in text generation. It assesses the semantic similarity between the model’s output and the target entity’s description. If the output closely matches the target while diverging from non-targets, it indicates that the model retains knowledge and has not effectively forgotten.
\subsubsection{Relearn Time.}
Relearn time refers to the number of training epochs required for a model to relearn and restore its performance on forgotten data. Studies~\cite{salun,i4,i6, zero} use a loss function to evaluate the model's performance on the target data and set a threshold below which the model's loss on the forgotten data points is considered sufficiently low. When the model's loss falls below this threshold, it is deemed to have successfully relearned the forgotten data. A higher number of training epochs generally indicates better forgetting performance by the model.

\subsubsection{Model Output.}
The performance of the model is largely reflected in its output. Some studies attempt to demonstrate the model's forgetting effect by measuring the difference in outputs between the model before and after forgetting. For example, some studies~\cite{i4,i6, bad,nguyen2020variational} use the distance between the outputs of two models for forgotten data or forgotten data features. \citet{ssse} proposed measuring unlearning effectiveness by calculating the $L_1$ distance between the confusion matrices of the original and unlearned models, and \citet{wang2025tape} audited unlearning by analyzing the posterior difference between the model outputs before and after unlearning on the same data. \citet{v_inf} quantified the residual forgotten information retained by the model by measuring the mutual information between the intermediate layer outputs and the labels of the data to be forgotten. Recently,  \citet{v_ai} proposed an unlearning verification method that leverages explainability tools to assess whether a model retains memory of target information. By analyzing differences in the model's responses before and after unlearning, the method uses attribution techniques to generate heatmaps based on intermediate activations and model outputs, thereby quantifying changes in the model's attention to key input regions.

\subsubsection {Model Sensitivity.}
Model sensitivity refers to the degree to which a model's output behavior or parameter gradients change when exposed to a given input, thereby reflecting the model’s dependency on that input.
This concept is concretely modeled in some works to enable unlearning verification.
\citet{v_fish} proposed a sensitivity-based evaluation method grounded in Fisher information to assess the effectiveness of machine unlearning. Specifically, the trace of the Fisher Information Matrix is used as a core metric to quantify the average gradient response of model parameters to the target data. A higher sensitivity indicates that the model has stably encoded the target data, whereas a successful unlearning process should reduce the Fisher information, leading to decreased sensitivity and increased epistemic uncertainty. This method assumes a white-box setting, as it requires access to gradient information, and supports efficient estimation through gradient-based approximations and computable upper bounds.
In contrast, \citet{zhou2025truvrf} proposed the TruVRF framework, which adopts an alternative definition of model sensitivity by measuring the magnitude of parameter updates during fine-tuning. The core idea is to infer whether the model retains memory of the target data based on how much its parameters need to adjust. If the model has not seen the data before, or has successfully forgotten it, fine-tuning will induce substantial parameter changes, resulting in high sensitivity. In contrast, if the model remembers the data, only minimal updates are required, corresponding to low sensitivity. This method also assumes a white-box setting, as it relies on accessing parameter-level changes during fine-tuning.
Although these methods differ in how they define and interpret sensitivity, they share a common principle: modeling the response strength of model parameters to input data as a proxy for memory retention. It is important to note that all such approaches inherently assume white-box access due to their reliance on internal parameter visibility, which may limit their applicability in certain real-world scenarios.

\section{Parametric Unlearning Verification}
\label{sec:Formalized}

\begin{table}[htbp]
\centering
\caption{Key symbols newly introduced for parametric unlearning verification.}
\label{tab:notation-cert}
\scalebox{0.8}{
\begin{tabular}{ll}
\toprule
\textbf{Symbol} & \textbf{Description} \\
\midrule
$\mathcal{D}'$ & Adjacent dataset differing from $\mathcal{D}$ in one data point \\
$N(\mathbf x, y)$ & Nearest retained sample to $(\mathbf x, y)$ from the same class\\
$\mathbf x$ & A single data point to be removed from the dataset \\
$S \subseteq \mathcal{H}$ & A measurable subset of hypotheses \\
$\epsilon$ & A privacy/forgetting parameter that bounds tolerated deviation (smaller $\epsilon$ implies stronger guarantee) \\
$\delta$ & Probability of failure \\
$\mathbf w_t$ & Model parameters at training step $t$ \\
$g_t$ & Update function applied at step $t$ \\
${d}_t$ & Training data used at step $t$ \\
$\tilde{d}^{(t)}$ & Replacement mini-batch excluding any forgotten data \\   
$(\mathbf w_i, \mathbf x_i, g_i)$ & Training log tuple\\
\bottomrule
\end{tabular}
}
\end{table}

\chdeleted{Behavioral verification methods begin by defining a target behavior to be evaluated and then assess whether the behavior of the unlearned model sufficiently resembles a reference behavior. In contrast, parametric verification methods operate at the model parameter level, verifying unlearning by analyzing the difference between the parameters of the unlearned model and those of a model that has never seen the forgotten data.}\chadded{Parametric unlearning verification methods are conducted through parameter-level analyses.} To support these analyses, several new symbols are introduced, as summarized in Table~\ref{tab:notation-cert}. Existing parametric approaches can be broadly categorized into two types: differential privacy verification and reproducibility verification. Table~\ref{tab:formal-verification-theory} provides a comparison of parametric verification methods.
\begin{table}[h]
\centering
\caption{Comparison of parametric verification methods.}
\label{tab:formal-verification-theory}
\renewcommand{\arraystretch}{1.3}
\scalebox{0.9}{
\begin{tabular}{@{}p{3.8cm}p{5.4cm}p{5.4cm}@{}}
\toprule
\textbf{Methods} & \textbf{Differential Privacy  Verification} & \textbf{Reproducibility Verification} \\
\midrule

\textbf{Core Dependency}
& Based on the indistinguishability of model parameters under neighboring datasets, formalized via differential privacy.
& Based on the ability to reconstruct the training trajectory from recorded updates. \\
\midrule

\textbf{Verification Logic} 
& Compares model distributions trained with and without \(D_f\). 
& Replays logs to ensure reproducibility and confirm absence of forgotten data. \\
\midrule

\textbf{Assumptions} 
& Often requires additional conditions on the loss function.
& Requires deterministic updates and full access to training logs. \\
\bottomrule
\end{tabular}
}
\end{table}

\subsection{Differential Privacy Verification}
Differential Privacy (DP) ~\cite{dp} is a data privacy protection mechanism that provides strong privacy guarantees when releasing statistical information,  while limiting the impact of individual data points on the results. The core idea is that even if a sample is added or removed,  the algorithm’s output probability distribution remains nearly unchanged,  thus preventing attackers from inferring whether a specific data point is present. Its mathematical definition is as follows:
\begin{definition}[\(\epsilon\)-Differential Privacy~\cite{dp}]
Let \( D \) and \( D' \) be two adjacent datasets (differing in at most one data point),  and let \( M \) be a randomized learning that maps data to a hypothesis space \( H \). For any subset of hypotheses \( S \subseteq H \),  we say that \( M \) satisfies \(\epsilon\)-Differential Privacy if:
\[
P(M(D) \in S) \leq e^{\epsilon} \cdot P(M(D') \in S)
\]
where \( \epsilon > 0 \) is the privacy parameter. A smaller \( \epsilon \) implies stronger privacy guarantees.
\end{definition}

By further relaxing the conditions,  the following holds:

\begin{definition}[(\(\epsilon,  \delta\))-Differential Privacy~\cite{dp}]
Under the same setting as above,  \( M \) satisfies \((\epsilon,  \delta)\)-Differential Privacy if:
\[
P(M(D) \in S) \leq e^{\epsilon} \cdot P(M(D') \in S) + \delta
\]
for all measurable subsets \( S \subseteq H\),  where \( \delta \) allows for a small probability of privacy failure.
\end{definition}
Inspired by the concept of differential privacy,  studies~\cite{guo, making} introduce a probabilistic formalization that limits the difference between the model weights returned by the machine unlearning and retraining algorithms,  providing provable guarantees of forgetting for the removed data. The following definitions are proposed:
\begin{definition}[\(\epsilon\)-Certified Forgetting~\cite{guo}]
Let \( A: D \to H \) be a randomized learning algorithm,  where the training dataset \( D \) induces a probability distribution over the hypothesis space \( H \). Define a data removal mechanism \( M \),  which aims to remove the influence of a sample \( \mathbf x \in D \) from the output of \( A(D) \). If,  for all possible training datasets \( D \subseteq X \),  all possible removal samples \( \mathbf x \in D \),  and all possible hypothesis sets \( S \subseteq H \),  the following inequality holds:  
\[
e^{-\epsilon} \leq \frac{P(M(A(D),  D,  \mathbf x) \in S)}{P(A(D \setminus \mathbf x) \in S)} \leq e^{\epsilon}
\]
then \( M \) is said to achieve \(\epsilon\)-Certified Forgetting. 
\end{definition}
This property ensures that the model distribution after removing \( \mathbf x \) is close to the model distribution that has never seen \( \mathbf x \),  making it difficult for an adversary to distinguish whether \( \mathbf x \) was used during training.
Guo et al.~\cite{guo} further relax the conditions and introduced the following result:

\begin{definition}[\((\epsilon,  \delta)\)-Certified Forgetting~\cite{guo}]
For \( \delta > 0 \),  if the data removal mechanism \( M \) satisfies the following relaxed conditions:
\begin{IEEEeqnarray}{rCl}
& P(M(A(D),  D,  \mathbf x) \in S) \leq e^{\epsilon} P(A(D \setminus \mathbf x) \in S) + \delta, \nonumber\\
& P(A(D \setminus \mathbf x) \in S) \leq e^{\epsilon} P(M(A(D),  D,  \mathbf x) \in S) + \delta, \nonumber
\end{IEEEeqnarray}
then \( M \) is said to achieve \((\epsilon,  \delta)\)-Certified Forgetting.  
\end{definition}

Here,  \( \delta \) controls the upper bound on the probability of failure,  allowing for a small divergence probability,  thereby relaxing the strict constraints of \(\epsilon\)-Certified Forgetting.
The above definition concerns differential privacy guarantees for individual data points.
To generalize certified forgetting to the multi-point deletion setting, several works \cite{guo, second2} have proposed treating the set of data to be removed as a deletion unit.
Building on this idea, subsequent studies \cite{vfl, vfr} have extended certified unlearning to the federated learning setting by enabling client-level unlearning, where a client’s entire dataset is treated as a single unit of deletion in the unlearning process.

It can be observed that this class of methods for achieving certified forgetting inherently has these verification capabilities within the model’s parameter space. We refer to this line of work as differential privacy unlearning verification. Several methods have been proposed in this context, which can be broadly categorized into first-order approaches based on function gradients and second-order approaches that leverage Hessian information or its approximations, as outlined below.

\subsubsection{First-order methods.}
First-order methods rely on gradient-related information to achieve provable unlearning. For example, \citet{langevin} combined differential privacy and proposed an approximate unlearning framework based on projection noise gradient descent. This framework is suitable for empirical risk minimization problems with smooth objectives,  providing provable forgetting capabilities.
\citet{chien2} proposed an approximate unlearning method based on projected noisy stochastic gradient descent,  offering certified unlearning guarantees by tracking the Wasserstein distance of the parameter distribution. This approach extends to multiple unlearning requests,  including sequential and batch settings,  and achieves better loss bounds through tighter analysis of the \( W_{\infty} \) distance. Additionally,  \citet{vfl} introduced provable forgetting in federated learning scenarios by adding Gaussian noise to the gradient during both model training and unlearning,  and applying a specific gradient ascent algorithm during the unlearning update. \citet{vfr} proposed the CFRU algorithm for provable forgetting in federated recommendation systems. This method rolls back and removes the target client’s historical updates,  uses a sampling strategy to reduce storage requirements,  estimates the deletion bias using Lipschitz conditions,  and compensates for model errors through an iterative scheme without the need for additional training. \citet{vword} provided the first theoretical guarantee for learning in pre-training and fine-tuning paradigms,  focusing on bag-of-words language models,  which can be applied to downstream tasks such as retrieval and classification. 

The above methods focus on convex or strongly convex functions. Recently, \citet{rewind} proposed a provable unlearning method for non-convex functions. This method performs unlearning by "rewinding" to earlier steps in the learning process before applying gradient descent on the loss function for retained data points, achieving a better balance between privacy,  utility,  and computational complexity. The method is shown to provide generalization guarantees for non-convex functions that satisfy the Polyak-Lojasiewicz inequality.

\subsubsection{Second-order methods.} These methods rely on Hessian or approximations of Hessian-related information to provide certified unlearning. Guo et al.~\cite{guo} first introduced a provable deletion mechanism for $L_2$-regularized linear models,  applicable to models trained with various convex loss functions (e.g.,  logistic regression). This mechanism significantly reduces the impact of deleted data points by applying a Newton step to the model parameters,  causing the residual error to decrease quadratically with the size of the training set. To better prove forgetting,  they further introduce a method based on random perturbation of the training loss to mask these residual errors.
This method requires retraining the model,  and to reduce computational costs,  \citet{remember} proposed an algorithm that enables sample deletion without accessing the original training data. The algorithm estimates the impact of deleted data on the model using the Hessian matrix and adjusts through noise perturbation,  ensuring provable forgetting. Recently,  \citet{FedRemoval} introduced a certified unlearning strategy for federated learning. This strategy performs unlearning using Newton updates on the server side,  simplifying the training process with a linear loss function,  requiring no client participation or additional storage. This method effectively controls the difference between the unlearned weights and the original optimal weights,  enabling efficient and certified unlearning.

Although the above methods provide proofs for forgetting,  they are not practical. They only consider convexity assumptions,  involve costly Hessian matrix computations,  and assume the learning model is in the empirical risk minimization setting. To enhance practicality in deep learning scenarios,  \citet{second2} investigated the problem of certified forgetting without relying on convexity assumptions. The study shows that by making simple modifications to Newton updates,  an approximation error bound for non-convex functions can be established,  and various approximation unlearning strategies for convex models can be adapted to improve their robustness in deep neural networks. Additionally,  to improve efficiency,  they propose a computationally efficient method to estimate the inverse Hessian in Newton updates and prove that this method retains bounded approximation errors.
\citet{hessianfree} introduced an innovative Hessian-free second-order unlearning method designed to overcome the limitations of traditional second-order algorithms,  such as high Hessian computation costs,  inability to handle multiple deletion requests online,  and reliance on convexity assumptions. By analyzing each sample’s impact on the training trajectory,  this method avoids explicitly computing the Hessian matrix,  enabling efficient online deletion request handling. Compared to existing methods,  this approach does not require the learning model to be an empirical risk minimizer and offers better scalability. Furthermore,  \citet{minmax} developed a new certified machine unlearning algorithm for min-max models,  achieving min-max unlearning steps with fully Newton-based updates using Hessian and incorporating Gaussian noise through differential privacy. To ensure certified unlearning,  this method injects calibrated Gaussian noise by carefully analyzing the similarity between min-max unlearning variables and retraining variables,  applicable to (strongly-)convex-(strongly-)concave loss functions.

Second-order forgetting methods have not only been widely explored in deep learning but have also made significant progress in graph neural networks (GNNs). \citet{efficient_graph} introduced the first proven theoretical guarantee for approximate graph unlearning,  focusing on addressing the issue of deletion requests for graph-structured data in GNNs. This method handles three types of unlearning requests: node feature unlearning,  edge unlearning,  and node unlearning. Through theoretical analysis of simple graph convolution and its generalized PageRank extension,  the study provides a theoretical guarantee for approximate unlearning. \citet{graph_2} effectively addressed unlearning requests by modeling the intermediate state of the optimization objective between instances with and without the data to be unlearned (such as nodes,  edges,  and attributes) and demonstrates a tighter bound in approximating actual GNN parameters.
Recently,  focusing on edge unlearning in graphs,  \citet{certified_edge} introduced a new influence function that efficiently computes the required updates while accounting for the neighborhood of deleted edges. This method removes the requested edges from the GNN without requiring retraining,  while also providing a certified unlearning guarantee.
Nevertheless,  these certified graph unlearning methods still require bounded model errors on precise node embeddings to maintain their certification guarantees. To address this challenge,  \citet{scalable} proposed ScaleGUN,  the first method to extend certified graph unlearning to billion-edge graphs. ScaleGUN integrates approximate graph propagation techniques into certified graph unlearning,  offering certification guarantees for three unlearning scenarios: node feature,  edge,  and node unlearning. 

\subsection{Reproducibility Verification}
\subsubsection{Proof-of-Unlearning (PoUL)}
\citet{poul} proposed a reproducible unlearning verification method inspired by the Proof-of-Learning (PoL)~\cite{pol}. PoL aims to verify whether a machine learning model has undergone effective training. It does this by logging the training process,  tracking model checkpoints,  the data points used,  and the hyperparameters associated with updates. These logs are typically represented as a series of tuples,  such as $\{(\mathbf w_i,  \mathbf x_i,  g_i)\}$, 
where $\mathbf w_i$ represents the model weights,  $\mathbf x_i$ represents the data point,  and $g_i$ represents the update rule. Auditors use these logs to reproduce checkpoints and compute the error between the actual and reproduced checkpoints. If the error is below a certain threshold,  the update is considered valid,  ensuring the correctness of the training process.

Expanding PoL to validate the correctness of "forgetting" specific data points,  \citet{poul} introduced the concept of PoUL,  which ensures that certain data points are successfully excluded during retraining. PoUL uses PoL logs to prove that the data points marked for "forgetting" were not used during model updates. \citet{fragile} identified two essential conditions for PoUL: reproducibility and removability.

\textbf{Reproducibility}: This requires that each model update step can be reproduced based on the update function,  and the error is within an acceptable tolerance,  expressed as:
\[
\|\mathbf w_t - g_t(\mathbf w_{t-1},  d_t)\| \leq \epsilon
\]
where $\mathbf w_t$ is the model parameter at step $t$,  $g_t$ is the update function,  $d_t$ is the data used for the step,  and $\epsilon$ is the tolerated error.

\textbf{Removability}: This ensures that the data points to be "forgotten" were not used in retraining,  expressed as:
\[
d_t \cap D_f = \emptyset
\]
where $d_t$ is the dataset at step $t$,  and $D_f$ is the set of data to be removed. Auditors verify the correctness of the "forget" operation by checking whether PoUL meets these two conditions.
These methods required recording and storing model parameters at every update step, which introduced significant storage overhead. To address this issue, some studies focused only on the final step of the unlearning process. For example, \citet{deltagrad} determined whether a predefined structural constraint was satisfied by comparing the parameters of a model retrained on $D_r$ with those of the target model. \citet{nguyen2022markov} further improved the distance metric by using the Frobenius norm of the difference between the corresponding weight matrices as a quantitative measure. These approaches did not rely on additional empirical behavior-level observation modules, but instead provided a formal verification pathway by directly measuring parameter-level deviations.

Although PoUL enables auditors to reproduce the unlearning process,  it still faces the risk of forgery. Attackers can manipulate logs by swapping data points,  generating seemingly valid records that pass verification~\cite{fragile,poul}. This vulnerability arises because PoUL grants model owners full access to training logs,  allowing them to alter specific steps without retraining the model.  

\subsubsection{PoUL with Cryptography.} To mitigate these risks,  \citet{tee} proposed a solution based on Trusted Execution Environments (TEEs) tailored for SISA unlearning scenarios. \chadded{TEE technology provides a protected, processor-supported execution environment that isolates critical code and data processing, thereby improving runtime integrity and reducing the risk of tampering with internal states and audit records.} \chreplaced{Specifically, t}
{T}his approach leverages Intel's Software Guard Extensions (SGX) to ensure the integrity of data tracking and model training. The solution consists of two layers: the authentication layer,  which tracks the impact of data on submodels and updates them accordingly after data deletion,  and the proof layer,  which guarantees the correctness of both the learning and inference processes by verifying that the designated data has been properly removed.  

Additionally,  \citet{crypt} introduced a cryptographic verifiable machine unlearning method that employs \chreplaced{Succinct Non-Interactive Arguments of Knowledge (SNARKs)}{SNARKs} and hash chains for verifiable computation and proof generation. \chadded{SNARKs provide succinct cryptographic proofs of computational correctness, allowing a verifier to ascertain protocol compliance without re-executing the full training procedure. Hash chains serve as tamper-evident log integrity mechanisms based on cryptographic hash functions, preventing retrospective modification or removal of recorded entries.} This framework is compatible with various unlearning techniques and has been validated through linear regression,  logistic regression,  and neural networks. The protocol ensures that users can verify the correctness of the unlearning process while maintaining model security and integrity.  

While these TEE- and cryptography-based methods enhance the reliability of unlearning verification,  their high computational overhead presents a significant challenge for practical deployment.

\section{Advantages and Limitations}  

\label{sec:Advantages_limitations}
\subsection{Key Dimensions for Evaluation.}
To systematically assess the effectiveness and practicality of unlearning verification methods, we introduce a set of key evaluation dimensions. These dimensions are derived from common challenges in verification design, practical deployment needs, and observed limitations in current research. They provide a structured lens through which to compare different methods across both behavioral and parametric categories.

\begin{itemize}[leftmargin=*, label={}]
\item \textit{Q1. Theoretical Guarantee:} Does the method provide formal, provable guarantees of forgetting?

\item \textit{Q2. Access Requirement:} Can the method operate in a black-box setting, where internal model parameters are inaccessible?

\item \textit{Q3. Sample-Level Verification:} Can the method evaluate forgetting at the level of individual samples?

\item \textit{Q4. Verification Accuracy:} Does the method offer high verification accuracy?

\item \textit{Q5. No Pre-Injected Data Required:} Is the method effective without requiring specially crafted or pre-injected data during training?

\item \textit{Q6. Efficiency and Scalability:} Is the method computationally efficient\chadded{, storage-efficient,} and scalable to large models?

\item \textit{Q7. Method Specificity:} Is the method general-purpose, or is it tailored to specific types of unlearning algorithms?
\end{itemize}

These dimensions form the basis of our comparative analysis in the following subsections. Each verification approach, whether behavioral or parametric, is evaluated against these criteria to highlight its strengths, limitations, and suitable application scenarios.

To facilitate comparison across different categories of methods, we use symbolic indicators to represent the level of support for each criterion:
\begin{itemize}
  \item[\cmark:] The methods in this category fully satisfy the requirement.
  \item[\hmark:] The majority of methods in this category satisfy the requirement.
  \item[\wmark:] Others.
  \item[\xmark:] The methods in this category do not satisfy the requirement.
\end{itemize}

\subsection{Behavioral Unlearning Verification}
Behavioral unlearning verification relies on analyzing the model's responses to specific data. This section explores the strengths and limitations of different behavioral verification methods.
\subsubsection{Fingerprint Verification}
Fingerprint verification methods, including passive fingerprint and active fingerprint.

\textit{Q1 Theoretical Guarantee} and \textit{Q2 Access Requirement}. 
All existing fingerprint verification methods lack formal theoretical guarantees regarding the effectiveness of the unlearning process. Consequently, they are inherently empirical in nature, and their results can only serve as indicative evidence rather than providing definitive assurance (\textit{{For All: Q1. \xmark}}).
As for access requirements, all existing {passive fingerprint} methods require white-box access (\textit{Q2. \xmark}). In contrast, most {active fingerprint} methods support black-box verification, with the exception of \citet{tam2024towards}, which operates under a white-box setting (\textit{Q2. \hmark}).

\textit{Q3 Sample-Level Verification.} 
Active fingerprint verification typically evaluates forgetting effectiveness by analyzing the collective activation patterns of a group of embedded samples, which makes it difficult to determine whether a specific data point has been successfully unlearned. To address this limitation, \citet{xu2024really} introduced a backdoor-based verification mechanism that constructs dedicated sample pairs for each target data point, thereby enabling sample-level unlearning verification. Although this approach provides fine-grained verification capabilities, it relies on uploading a substantial amount of non-forgettable data, which may complicate its integration into privacy-sensitive or constrained application settings (\wmark).

Passive fingerprint verification, on the other hand, has been shown to support single-sample-level unlearning verification in classification models~\cite{xuan2025verifying}. However, in generative models, existing studies~\cite{xuan2025verifying, zhang2024generate} do not assess whether a specific data point has been forgotten, but instead focus on evaluating the model’s ability to forget an entire semantic concept (\wmark).

\textit{Q4 Verification Accuracy.} The accuracy of fingerprint verification methods largely depends on factors such as the quality of the injected fingerprints and the specifics of the model's training process (\textit{For All: Q4. \xmark}). For example, active fingerprint methods can be sensitive to changes in model architecture or security defenses (e.g., backdoor~\cite{wang2024eclipse,yu2025backdoor,zhou2025darkhash,wan2025mars} removal techniques). In contrast, passive fingerprint methods often rely on adversarial examples~\cite{lu2026pretrain, yu2025towards, yu2022towards, lu2026robots}, whose effectiveness may significantly degrade in the presence of adversarial defense strategies, thereby undermining the reliability of the verification.

\textit{Q5 No Pre-Injected Data Required.} 
Active fingerprint methods rely heavily on pre-injected data (\textit{Q5. \xmark}). This means that special samples must be embedded during training to enable subsequent verification, which limits their applicability in scenarios where post-hoc verification on arbitrary user data is required. In contrast, passive fingerprint methods offer greater flexibility, as they typically support verification through post-training queries without requiring any intervention during the training process (\textit{Q5. \cmark}).

\textit{Q6 Efficiency and Scalability} and \textit{Q7 Method Specificity.}
Fingerprint verification methods are generally efficient\chreplaced{, require minimal storage and highly scalable.}{ and highly scalable.} Since they do not require retraining the model, these methods can be deployed quickly and are applicable to large-scale models and datasets (\textit{For All: Q6. \cmark}). Moreover, fingerprint does not depend on any specific unlearning strategy, making it broadly compatible with a wide range of unlearning approaches. This generality enables fingerprint methods to maintain strong adaptability and practical utility across diverse application scenarios (\textit{For All: Q7. \cmark}).

\subsubsection{Privacy Analysis Verification}
Privacy analysis verification methods can be broadly categorized into data inference and data reconstruction approaches.

\textit{Q1 Theoretical Guarantee} and \textit{Q2 Access Requirement}.
Privacy analysis verification methods lack formal theoretical guarantees for the unlearning process, as their validation relies primarily on empirical evaluations via privacy analysis and does not offer rigorous theoretical support (\textit{Q1. \xmark}).

In terms of access requirements, privacy analysis methods exhibit a range of access assumptions. For example, data reconstruction techniques vary in their requirements, with some relying on black-box access \cite{graves2021amnesiac,xu2024evaluating} and others on white-box access~\cite{,bertran2024reconstruction,hu2024learn}~(\textit{Q2. \wmark}). Mainstream data inference methods also frequently operate under black-box settings, although recent studies~\cite{wang2024towards} proposed white-box variants to improve inference accuracy, thus they are rated as \textit{\hmark} in Q2. 


\textit{Q3 Sample-Level Verification} and \textit{Q4 Verification Accuracy}.
Data inference methods enable sample-level verification by assessing whether the target model retains privacy-relevant information associated with specific data samples (\textit{Q3.  \cmark}). While most data reconstruction methods are capable of recovering information at the sample level, the study~\cite{graves2021amnesiac} only supports class-level reconstruction and lacks the granularity required for single-sample verification  (\textit{Q3. \hmark}).

In addition, the accuracy of these methods largely depends on factors such as model architecture, training dynamics, and the effectiveness of the attack design. For instance, the quality of the attack model, shifts in the training data distribution, and model updates can all affect the stability of the verification results. Due to these uncertainties, the verification accuracy of privacy analysis methods is generally unstable (\textit{Q4.  \xmark}).

\textit{Q5 No Pre-Injected Data Required}, \textit{Q6 Efficiency and Scalability}, and \textit{Q7 Method Specificity}. Privacy analysis verification methods generally do not require pre-injected special samples (Q5. \cmark), making them more flexible and applicable for post-hoc evaluations. However, computational efficiency varies depending on the attack type used. Data reconstruction methods do not require retraining the model and rely on optimization or analytical techniques, making them relatively lightweight with good efficiency and scalability (Q6. \cmark ). In contrast, many data inference methods typically require training multiple auxiliary models, which can incur significant computational overhead \chadded{and storage overhead} in large-scale applications.
Recently, \citet{wang2024towards} demonstrated that inference verification can be performed without relying on auxiliary models by directly leveraging the unlearned model itself, thereby improving efficiency to some extent (Q6. \wmark). Additionally, privacy analysis methods are broadly applicable and can be used to assess various types of unlearning processes, so they score \textit{\cmark} on \textit{Q7}.

\subsubsection{Model Performance Verification.}
Model performance verification focuses on evaluating aspects such as accuracy, relearn time, model output, and model sensitivity.

\textit{Q1 Theoretical Guarantee} and \textit{Q2 Access Requirement}.
Model performance serves as an intuitive indicator but does not offer formal theoretical guarantees for the unlearning process, and is thus rated as (\textit{Q1. \xmark}).
In terms of access requirements, methods based on accuracy and relearn time typically require only black-box access, without the need for internal model parameters or training data, which greatly enhances their flexibility and applicability (\textit{Q2. \cmark}).
Output methods also mostly relies on black-box outputs, although some studies~\cite{v_ai,v_inf} suggested that using intermediate layer outputs can yield more effective verification results (\textit{Q2. \hmark}).
In contrast, model sensitivity methods depend on a white-box setting, as they require access to model parameters to compute gradients (\textit{Q2. \xmark}).

\textit{Q3 Sample-Level Verification} and \textit{Q4 Verification Accuracy}.
Methods based on task quality and relearn time generally focus on overall performance on the unlearned dataset and cannot directly evaluate individual samples (\textit{Q3. \xmark}).
In comparison, model output and model sensitivity methods assess the model's response to specific samples and can thus provide more fine-grained, sample-level verification (\textit{Q3. \cmark}).
However, since these methods rely on model behavior, they are sensitive to factors such as data distribution shifts and hyperparameter tuning, leading to performance fluctuations and making their verification accuracy and stability unreliable (\textit{Q4. \xmark}).

\textit{Q5 No Pre-Injected Data Required}, \textit{Q6 Efficiency and Scalability}, and \textit{Q7 Method Specificity}.
Model performance methods do not rely on pre-injected data for verification (Q5. \cmark). Methods based on relearn time, model output, and model sensitivity are efficient\chadded{, low-storage-overhead,} and scalable, as they do not require retraining the entire model. This makes them suitable for large models and datasets (Q6. \cmark). For task quality, most approaches allow efficient evaluation by measuring how well the model completes the task. Recently, \citet{nguyen2022markov} proposed a method that retrains a separate model and compares its task quality with that of the unlearned model, which results in significant computational overhead \chadded{and storage overhead} (Q6. \hmark). As for method specificity, performance methods apply to various unlearning mechanisms (Q7. \cmark).

\subsection{Parametric Unlearning Verification}
Parametric verification relies on theoretical proofs and mathematical guarantees to assess whether a model has truly forgotten specific data. This section explores the strengths and weaknesses of different parametric verification methods.
\subsubsection{Differential Privacy Verification}
The differential privacy verification is a validation mechanism that provides strong theoretical proof, ensuring that after data deletion, the model cannot leak the existence of the removed data points.

\textit{Q1 Theoretical Guarantee}.
Differential privacy verification methods are supported by a strong theoretical foundation, based on rigorous mathematical definitions. These methods provide quantifiable guarantees for the forgetting process through mathematical proofs and privacy protection algorithms (Q1. \cmark). 

\textit{Q2 Access Requirement}.
Differential privacy methods typically rely on white-box access to the model's internal parameters. This is because to verify whether the model satisfies differential privacy conditions, auditors must have access to the model’s parameters  (Q2. \xmark). 

\textit{Q3 Sample-Level Verification} and \textit{Q4 Verification Accuracy}.
Differential privacy methods offer fine-grained control over each data point, enabling sample-level verification  (Q3. \cmark).  Additionally, since they are based on solid theoretical guarantees, these methods are more trustworthy than behavioral approaches  (Q4. \cmark). 

\textit{Q5 No Pre-Injected Data Required}, \textit{Q6 Efficiency and Scalability} and \textit{Q7 Method Specificity}.
Differential privacy methods do not rely on pre-injected specific samples; instead, the verification process only requires the use of existing data (Q5. \cmark). While these methods offer strong theoretical guarantees, their computational efficiency and scalability are limited. Specifically, first-order methods necessitate the verification of each model parameter and gradient, whereas second-order methods involve costly Hessian matrix calculations, resulting in significant computational overhead. Additionally, \chadded{these methods typically require retaining extra training-time information, such as model checkpoints and historical updates, leading to considerable storage overhead.} \chadded{They also often}\chdeleted{most of these methods} rely on strong assumptions, such as convexity or strong convexity, which further restrict their scalability (Q6. \xmark).

Moreover, differential privacy methods can only enable effective unlearning verification when used in conjunction with their corresponding training mechanisms or algorithmic frameworks, which limits their applicability in certain unlearning scenarios (\textit{Q7. \xmark}).

 \subsubsection{Reproducible Unlearning Verification}
Reproducible unlearning verification methods such as PoUL and its cryptographic variants aim to provide rigorous and verifiable guarantees that a model has genuinely forgotten specific data points during the unlearning process.

\textit{Q1 Theoretical Guarantee \& Q2 Access Requirement}.  
These methods offer strong theoretical guarantees by ensuring that model updates and checkpoints can be validated against expected changes, making the verification process theoretically sound ({Q1.  \cmark}). However, they typically require access to internal model parameters, weights, and training logs—resources often unavailable in commercial settings—thus limiting their practical applicability ({Q2. \xmark}).

\textit{Q3 Sample-Level Verification \& Q4 Verification Accuracy}.  
Reproducible methods support fine-grained verification at the individual sample level, ensuring that forgotten data is excluded during retraining ({Q3. \cmark}). PoUL achieves high verification accuracy through its repeatability and precise tracking of model updates ({Q4. \cmark}). Cryptographic extensions, such as those using Trusted Execution Environments (TEEs) or SNARKs, further enhance trust by providing tamper-proof proofs of unlearning ({Q4.  \cmark}).

{\textit{Q5 No Pre-Injected Data Required, Q6 Efficiency and Scalability, \& Q7 Method Specificity}.}  
These methods do not rely on specially crafted or pre-injected data; as long as relevant training and update records are maintained, they can be applied across various unlearning scenarios with flexibility ({Q5. \cmark}). Nonetheless, reproducing model checkpoints and verifying updates can be computationally expensive \chadded{and can incur substantial storage overhead}, especially for large models or datasets, which limits scalability ({Q6. \xmark}). Moreover, while the PoUL method exhibits broad applicability across various unlearning algorithms ({Q7. \cmark}), its cryptographic variants, despite enhancing verification reliability, may reduce generality. For instance, the approach proposed in \cite{tee} is primarily limited to SISA-based settings ({Q7. \wmark}).

\begin{table}[htbp]
\centering
\caption{Comparison of unlearning verification methods.The table presents the evaluation of verification methods across seven key dimensions: \textbf{Q1-Q7}. Each method category is scored using the following symbols:
\cmark\ (fully satisfy), 
\hmark\ (mostly satisfies, with some caveats),
\wmark\ (Others), \xmark\ (does not satisfy).
The seven dimensions evaluated are as follows: 
\textbf{Q1. Theoretical Guarantee}, \textbf{Q2. Access Requirement}, \textbf{Q3. Sample-Level Verification}, \textbf{Q4. Verification Accuracy}, \textbf{Q5. No Pre-Injected Data Required}, \textbf{Q6. Efficiency and Scalability}, and \textbf{Q7. Method Specificity}.}
\label{tab:unlearning_verification_all}
\scalebox{0.95}{
\begin{tabular}{@{}llccccccc@{}}
\toprule
\textbf{Category} & \textbf{Method} & \textbf{Q1} & \textbf{Q2} & \textbf{Q3} & \textbf{Q4} & \textbf{Q5} & \textbf{Q6} & \textbf{Q7} \\
\midrule

\multirow{2}{*}{Fingerprint} 
& Passive Fingerprint & \xmark & \xmark & \wmark & \xmark & \cmark & \cmark & \cmark \\
& Active Fingerprint & \xmark & \hmark & \wmark & \xmark & \xmark & \cmark & \cmark \\

\midrule
\multirow{2}{*}{Privacy Analysis} 
& Data Reconstruction & \xmark & \wmark & \hmark & \xmark & \cmark & \cmark & \cmark \\
& Data Inference & \xmark & \hmark & \cmark & \xmark & \cmark & \wmark & \cmark \\

\midrule
\multirow{4}{*}{Model Performance} 
& Task Quality & \xmark & \cmark & \xmark & \xmark & \cmark & \hmark & \cmark \\
& Relearn Time & \xmark & \cmark & \xmark & \xmark & \cmark & \cmark & \cmark \\
& Model Output & \xmark & \hmark & \cmark & \xmark & \cmark & \cmark & \cmark \\
& Model Sensitivity & \xmark & \cmark & \xmark & \xmark & \cmark & \cmark & \cmark \\

\midrule
\multirow{2}{*}{Differential Privacy Verification} 
& First-order Methods & \cmark & \xmark & \cmark & \cmark & \cmark & \xmark & \xmark \\
& Second-order Methods & \cmark & \xmark & \cmark & \cmark & \cmark & \xmark & \xmark \\

\midrule
\multirow{2}{*}{Reproducible Verification} 
& PoUL & \cmark & \xmark & \cmark & \cmark & \cmark & \xmark & \cmark \\
& PoUL with Cryptography & \cmark & \xmark & \cmark & \cmark & \cmark & \xmark & \wmark \\

\bottomrule
\end{tabular}
}
\end{table}
\subsection{Comparison with Different Verification}

Table~\ref{tab:unlearning_verification_all} presents a systematic comparison of existing unlearning verification methods, evaluated across the key dimensions introduced earlier (Q1 to Q7). Based on this foundation, we further analyze the respective strengths and limitations of behavioral and parametric approaches.

Behavioral methods are generally effective in practical deployment due to their low access requirements and high computational efficiency, making them particularly suitable for black-box environments such as commercial APIs. However, they lack formal guarantees, offer limited support for sample-level verification, and often exhibit instability in real-world scenarios. For instance, fingerprint techniques may fail to reliably activate the embedded signals when model architectures or data distributions shift, resulting in false positives or negatives. Privacy analysis methods also vary significantly across different models and tasks, with inference accuracy often degrading after model updates. Similarly, performance methods are sensitive to factors such as random initialization and hyperparameter tuning, leading to inconsistent results and reduced reproducibility.

In contrast, parametric verification methods provide provable guarantees and support fine-grained, sample-level evaluation. Nonetheless, they typically require full access to the model’s internal parameters, incur substantial computational \chadded{and storage} overhead, and often rely on specific assumptions or customized protocols designed for particular unlearning mechanisms.

This comparison highlights a core trade-off between practicality and rigor. While behavioral methods are easier to deploy and more amenable to external auditing, they lack theoretical soundness. Parametric methods, by contrast, offer strong guarantees but face limitations in scalability and general applicability. Addressing this gap through hybrid verification frameworks that combine practical accessibility with formal robustness presents a promising direction for future research.
\section{Threats to Verification Reliability}
\label{sec:threat}
\subsection{Threat Model}
We consider a setting where the unlearning protocol has been invoked, and an external verifier is tasked with determining whether a machine learning model has successfully forgotten a designated subset of data, denoted \(D_f\). Depending on the verification framework, the auditor may access the final model parameters, intermediate training logs, or behavioral outputs in response to specific queries.
Our threat model assumes that the {model provider may act dishonestly}. That is, the provider may claim to have unlearned \(D_f\) while the model still retains knowledge or influence from the supposedly forgotten data. We assume that the provider has full control over the training pipeline and can manipulate training data, modify update trajectories, alter audit logs, or fine-tune the final model to pass verification checks.

Under this adversarial assumption, the central research question becomes:
\begin{quote}
\textit{How can a dishonest model provider mislead the verifier into falsely believing that the model no longer retains any influence from the forgotten data?}
\end{quote}
Based on how the verifier evaluates the model, threats can be broadly categorized into two types: those that exploit the {parameter space}, and those that target the {behavioral space}. The following sections analyze how each category of threat undermines the reliability of unlearning verification mechanisms.

\subsection{Threats from the Parameter Space}
Parametric unlearning verification methods often rely on assessing similarity in the model’s parameter space---such as comparing final model weights or replaying training logs to verify whether forgotten data was used. However, recent studies have revealed that such mechanisms face fundamental security vulnerabilities when the model provider is considered untrusted. The core issue stems from the stochastic nature of modern training pipelines, which leads to a {non-unique mapping between training data and model parameters}, thereby enabling adversarial forgeries.

Thudi et al.~\cite{poul} introduced the notion of a \emph{forging map}, which allows an attacker to bypass verification by simulating the removal of an entire subset of data \(D_f\) without actually deleting it. 
Formally, let \(d^t\) denote the mini-batch used at training step \(t\), where \(d^t \subseteq D\). If \(d^t\) contains any forgotten data, i.e., \(d^t \cap D_f \neq \emptyset\), the attacker aims to find a replacement batch \(\tilde{d}^{(t)} \subseteq D \setminus D_f\) such that the resulting parameter update closely matches the original: 
\[
\left\| g(\mathbf w_i, d^t) - g(\mathbf w_i, \tilde{d}^{(t)}) \right\| \leq \varepsilon. 
\]
When \(\varepsilon\) is sufficiently small (empirically as low as \(10^{-6}\)), the forged unlearning record can easily pass verification. This strategy can also circumvent differential privacy verification, as the resulting parameter difference is negligible from the perspective of a model that has never encountered \(D_f\).

To further improve attack efficiency, Zhang et al.~\cite{fragile} proposed an \emph{input-space nearest neighbor substitution strategy}. Rather than minimizing the gradient deviation, this approach directly replaces each forgotten sample in a minibatch with the most similar retained sample from the same class, based on Euclidean distance. For each forgotten point \((x, y) \in D_f\), the attacker selects:
\[
N(\mathbf x, y) = \arg\min_{(\mathbf x^*, y^*) \in D \setminus D_f,\, y^* = y} \|\mathbf x^* - \mathbf x\|.
\]
This method avoids the cost of gradient alignment or retraining while preserving update similarity, significantly improving attack efficiency. Experiments show that the forged trajectories produced in this way can reliably bypass verification mechanisms under practical tolerance thresholds (e.g., \(\varepsilon = 10^{-3}\)).

In summary, the success of such attacks stems from the non-invertibility of the parameter space: the same parameter trajectory can be produced by different training datasets. As long as the model provider controls the training process, they can construct forged logs that exclude \(D_f\) while remaining verifiable under existing parametric approaches.

To mitigate these risks, some recent works~\cite{crypt, tee} proposed incorporating cryptographic enhancements into the PoUL framework, such as trusted execution environments or zero-knowledge proofs. While these techniques improve integrity and tamper-resistance, their significant computational and deployment overhead limits their practicality in large-scale or resource-constrained settings.

\subsection{Threats from the Behavioral Space}

Behavioral unlearning verification methods primarily rely on observable model behaviors, rather than on cryptographic or formally grounded guarantees. While this design facilitates deployment in black-box environments and offers practical engineering flexibility, it inherently lacks robustness and exposes a broad and weakly protected attack surface. Recent research~\cite{poul} has shown that, compared to parametric verification, behavioral approaches are more vulnerable to forgery attacks in the parameter space. However, the risks extend beyond parameter-level manipulation. Behavioral verification also faces credibility threats from {behavioral-space forgery}, where adversaries deliberately manipulate model behavior to appear compliant during audit, while covertly retaining information about the supposedly forgotten data.
We categorize the primary risks in behavioral-space verification as follows:
\begin{enumerate}
    \item{\textbf{Suppressing Behavioral Fingerprints.}}
Verification methods based on fingerprint detection rely on the model’s distinctive responses to crafted probes to determine whether remnants of forgotten data persist. However, such behavioral signatures can be neutralized. Adversaries may train the model adversarially to suppress fingerprint activation~\cite{adv1, adv2}, or apply input filtering and sanitization mechanisms to block suspicious fingerprint queries~\cite{adv3, adv4}. Targeted fingerprint removal techniques~\cite{backdoor1, watermark1} can eliminate verification triggers while preserving memorized content, thereby invalidating the behavioral cues relied upon by the auditor.

\item{\textbf{Obfuscating Privacy Leakage Signals.}}
Privacy analysis verification typically rely on specific privacy attacks to extract identifiable privacy signals associated with forgotten data, in order to assess whether such data continues to influence the model. However, these signals can be deliberately obfuscated. For example, retraining with differential privacy~\cite{mia_d1}, injecting noise into gradients, or clipping model outputs~\cite{mid1, mid2} can all reduce the effectiveness of such analysis. As a result, a model may appear to have forgotten certain data, but this effect may be due to defense mechanisms suppressing the adversary's ability to identify the forgotten data, rather than actual data removal.

\item{\textbf{Spoofing Performance Degradation.}}
Model performance methods monitor performance metrics to infer whether unlearning has occurred. Yet these indicators can be spoofed by adversaries. Through data poisoning~\cite{data_poison1, data_poison2}, the model’s performance can be manipulated to mimic the expected degradation or improvement patterns. In such cases, the model passes the performance check without performing actual data removal, undermining the verification’s trustworthiness.
\end{enumerate}

\medskip
In all cases, the root vulnerability lies in the {decoupling between observable behavior and true memory state}: behavioral signals can be engineered independently of actual forgetting. Without tighter coupling between external behavior and internal data provenance, behavioral verification remains vulnerable to strategic manipulation. This calls for the integration of behavioral modules with stronger formal or cryptographic guarantees to ensure robustness against behavioral forgery.

\section{Open Questions and Future Directions}
\label{sec:open}

\paragraph{Unified Definition of Forgetting.}  
A fundamental challenge in unlearning research is the absence of a unified and operational definition of forgetting. Current approaches rely on diverse criteria, including fingerprint \cite{zhang2024generate,tam2024towards,goel2022towards}, privacy analysis \cite{mia_game1, mia_game2, mia_game3}, and model performance \cite{graves2021amnesiac,acc2}. This definitional fragmentation creates a critical ambiguity: if a model passes certain verification protocols but fails others, can it still be considered to have truly forgotten the data? Without a common standard, service providers may selectively adopt favorable verification strategies and claim compliance without achieving the intended privacy guarantees. The lack of formal consensus not only impedes fair comparison across methods but also opens the door to unverifiable or misleading claims. Future research should aim to establish a rigorous, interpretable, and widely applicable definition of forgetting that accommodates various verification paradigms and practical deployment needs.

\paragraph{Reliable Unlearning Verification.}  
Beyond definitional ambiguity, the reliability of existing verification methods remains a pressing concern. Many behavioral approaches, such as those based on privacy analysis or fingerprint, depend heavily on model outputs and the effectiveness of auxiliary attack modules. When these modules are poorly calibrated or overly sensitive to distributional shifts, the resulting verification can become unstable or misleading.
At the same time, theoretically grounded methods, including differential privacy verification and reproducibility verification, also suffer from fundamental limitations. For instance, reproducibility verification assumes that if a model trained with and without a specific data point yields identical parameters, then the data has been successfully forgotten. However, as shown in~\cite{poul}, it is possible to construct datasets that differ by a single point \(\mathbf x\) yet result in nearly indistinguishable models. This allows for false claims of forgetting without any actual data removal.
Although recent works~\cite{tee,crypt} have explored the integration of cryptographic primitives into reproducibility verification to enhance its trustworthiness, these techniques remain computationally intensive and difficult to scale. This underscores the need for verification methods that combine theoretical rigor with practical feasibility. Such methods should provide reliable forgetting guarantees without imposing excessive computational or access requirements.

\paragraph{Limited Robustness Under Practical Constraints.}  
Despite growing interest in unlearning verification, many methods are developed under overly idealized assumptions. It is often presumed that data distributions are stationary~\cite{chen2021machine,salun}, model internals are fully accessible~\cite{guo,second2}, and the training process is entirely reproducible~\cite{poul,crypt}. These assumptions rarely hold in real-world deployments, where models are subject to stochastic training dynamics, evolving input distributions, and limited access due to privacy policies or API restrictions. Such constraints introduce substantial uncertainty that current methods are ill-prepared to handle. For instance, small perturbations in initialization or mini-batch order can significantly affect model outputs, undermining the reliability of behavior signals. Similarly, black-box environments make gradient-based or log-based verification infeasible. This gap between theoretical design and deployment reality remains one of the least addressed yet most impactful challenges. Without explicitly accounting for these constraints, verification results may become unstable, misleading, or fundamentally inapplicable. Future research must incorporate robustness into the core design of verification protocols, including tolerance to randomness, adaptability to shifting distributions, and operability under partial or noisy access. Otherwise, verification will remain a fragile component that functions only in controlled settings and silently fails in practical use.

\paragraph{Generalization Across Different Samples.}  
Another insufficiently explored limitation is the inconsistency of verification effectiveness across different data samples. Many behavioral
methods, particularly those based on privacy analysis or perform exceptionally well on atypical or poorly generalized samples, which are the most likely to be memorized by the model~\cite{long_mem1,long_mem2}. However, they often underestimate the forgetting failure rate on well-generalized samples, leading to incomplete or biased assessments. This inconsistency reflects a deeper issue: current verification techniques often implicitly assume that all samples are equally verifiable, ignoring the heterogeneous learning dynamics inherent to modern models. Without accounting for sample-specific factors such as gradient influence, model sensitivity, or generalization difficulty, these methods risk producing misleading evaluations of unlearning effectiveness. Future work should aim to develop verification strategies that support consistent and fair assessments across a broad spectrum of data instances.

\paragraph{Verification in the Duplicate Data.}  
Real-world datasets are rarely composed of unique or independent samples. Instead, they often contain duplicates or semantically similar instances originating from different users or data sources. In such cases, deleting a particular sample may not remove its semantic influence if similar data remains in the training set. This creates a critical blind spot for current verification methods: systems may technically fulfill a deletion request while still retaining equivalent information elsewhere, thus rendering the forgetting claim misleading~\cite{dupl}. Designing verification frameworks capable of detecting and quantifying residual memorization in the presence of redundancy or correlation is a significant open challenge. Future research should explore verification mechanisms with similarity-awareness or representation-level sensitivity to ensure that claims of forgetting reflect the model’s true information retention status.

\paragraph{\chadded{Hierarchical Hybrid Verification.}} \chadded{Existing unlearning verification methods generally involve a trade-off between practicality and rigor. Behavioral verification methods are typically easier to deploy, but the evidence they provide is relatively indirect. In contrast, parameter-based verification methods can provide stricter forgetting guarantees, but they typically incur higher computational and storage overhead. To address this limitation, a promising future direction is to develop a hierarchical hybrid verification framework. The core idea is to make staged verification decisions by jointly considering sample importance and verification cost: samples are first prioritized by importance, and low-importance unlearning samples are verified using low-cost behavioral methods; for high-value samples, parameter-based verification is then adaptively triggered to provide stronger evidence or additional review. This design can improve the credibility and specificity of unlearning verification while maintaining overall scalability.}
\paragraph{\chadded{Verification in Large Language Models.}} \chadded{The rapid rise and widespread deployment of large language models (LLMs) have intensified the need to protect the right to data erasure and, crucially, to verify whether unlearning has actually been achieved. However, due to the massive scale of these models and the high cost of training, existing methods are largely limited to behavior-level unlearning verification. Moreover, in the absence of an exact reference model obtained via full retraining, behavior-level evaluation lacks an alignable and calibratable ground truth, making its validity difficult to guarantee and limiting its ability to provide robust, auditable evidence. Future work should prioritize the development of {calibratable proxy ground truth} mechanisms, for example by constructing partially retrainable local references (e.g., retraining specific submodules or small localized subsets of data) and combining multiple independent proxy signals for cross-consistency checks. Such designs may enable more reliable, comparable, and verifiable unlearning evaluation even when full retraining-based references are infeasible.}

\section{Conclusion}
Machine unlearning is a process that enables machine learning models to forget specific training data through specialized techniques. While these techniques provide algorithm-level forgetting or approximate forgetting, the honesty of the service providers offering unlearning APIs often remains questionable. Therefore, verification becomes a crucial element to enhance the reliability of machine unlearning. However, existing works lack a comprehensive and thorough survey of this field. To fill this gap, we present the first survey on machine unlearning verification. We classify existing verification methods into behavioral and parametric types, analyzing their respective advantages and limitations. Additionally, we examine the associated vulnerabilities and threats and identify several open research questions. We believe that our work can advance the study of the reliability and security of machine unlearning, thus enabling machine learning systems to more reliably handle the forgetting of sensitive data.
  \bibliographystyle{ACM-Reference-Format}
    \bibliography{sample-base}
\end{document}